%% file: iclr2027_conference.tex
\documentclass{article} 
\usepackage{iclr2027_conference,times}

\usepackage[utf8]{inputenc} 
\usepackage[T1]{fontenc}    
\usepackage{hyperref}       
\usepackage{url}            
\usepackage{booktabs}       
\usepackage{amsfonts}       
\usepackage{nicefrac}       
\usepackage{microtype}      
\usepackage{xcolor}         
\usepackage{graphicx}
\usepackage{subcaption}
\usepackage{amsmath}
\usepackage{amssymb}
\usepackage{amsthm}
\usepackage{mathtools}
\usepackage{multirow}
\usepackage{tikz}
\usepackage{algorithm}
\usepackage{algorithmic}
\usepackage{enumerate}
\usepackage{wrapfig}
\usepackage[table]{xcolor}
\usepackage{mathtools}
\usepackage{tcolorbox}
\usepackage{enumitem}

\newcommand{\refig}[1]{Fig.~\ref{#1}}
\newcommand{\refapp}[1]{App.~\ref{#1}}
\newcommand{\refalg}[1]{Alg.~\ref{#1}}
\newcommand{\refsec}[1]{Sec.~\ref{#1}}

\newtcolorbox{hypothesisbox}[1]{colback=blue!5!white,colframe=blue!75!black,fonttitle=\bfseries,title=#1}

\hypersetup{
    colorlinks=true,
    citecolor=neublue,     
    linkcolor=red,      
    urlcolor=neublue       
}

\definecolor{neublue}{HTML}{3F51B5}     
\definecolor{neugreen}{HTML}{2E7D32}    
\definecolor{neumagenta}{HTML}{AD1457}  
\definecolor{neupurple}{HTML}{8E24AA}   
\definecolor{neuorange}{HTML}{E64A19}   
\definecolor{neugreenpastel}{HTML}{A5D6A7} 
\definecolor{neulavender}{HTML}{E9D5FF} 

\title{TSR: Trajectory‑Search Rollouts for Multi‑Turn RL of LLM Agents}

\author{%
  Aladin Djuhera$^1$, Swanand Kadhe$^2$, Farhan Ahmed$^2$, Syed Zawad$^2$, Heiko Ludwig$^2$, Holger Boche$^1$ \\
  $^1$ Technical University Munich \texttt{\{aladin.djuhera,boche\}@tum.de} \\
  $^2$ IBM Research \texttt{\{swanand.kadhe,farhan.ahmed,szawad,hludwig\}@ibm.com}
}

\iclrfinalcopy 

\begin{document}

\maketitle


\input{chapters/0_abstract}


\input{chapters/1_introduction}
\input{chapters/2_background}
\input{chapters/3_trajectory_search_rollouts}
\input{chapters/4_experimental_setup}
\input{chapters/5_results}
\input{chapters/6_conclusion}


\input{other/acknowledgements}


\bibliography{bibliography}
\bibliographystyle{unsrtnat}


\appendix
\input{appendix/A_related_work}
\input{appendix/B_policy_optimization}
\input{appendix/C_search_strategies}
\input{appendix/D_environments}
\input{appendix/E_training_setup}
\input{appendix/F_supplementary_results}
\input{appendix/G_limitations}


\end{document}

%% file: chapters/0_abstract.tex
\begin{abstract}
    Advances in large language models (LLMs) are driving a shift toward using reinforcement learning (RL) to train agents from iterative, multi-turn interactions across tasks.
    However, multi-turn RL remains challenging as rewards are often sparse or delayed, and environments can be stochastic.
    In this regime, naive trajectory sampling can hinder exploitation and induce mode collapse.
    We propose \textbf{TSR} (\textbf{T}rajectory-\textbf{S}earch \textbf{R}ollouts), a training-time approach that repurposes test-time scaling ideas for improved per-turn rollout generation.
    TSR performs lightweight tree-style search to construct higher-quality trajectories by selecting promising actions and trajectory prefixes during rollout generation.
    This improves rollout quality while preserving stable policy optimization and remains compatible with standard policy-gradient optimizers by design.
    Across Sokoban, FrozenLake, and WebShop, TSR achieves success-rate gains of up to 15 percentage points and converges in fewer optimization steps, while trading additional training-time rollout compute for stronger policies that require no search at inference time.
    By moving search from test time to the rollout stage of training, TSR provides a modular mechanism for stronger multi-turn agent learning, complementary to existing frameworks and rejection-sampling-style selection methods.
\end{abstract}

%% file: chapters/1_introduction.tex
\section{Introduction}
\label{sec:introduction}

Large language models (LLMs) are increasingly trained as interactive agents via reinforcement learning (RL), where the model acts over multiple turns in an environment and learns from feedback obtained through interaction rather than from static corpora \citep{nakano2022webgptbrowserassistedquestionansweringhuman, zhou2024archertraininglanguagemodel, zhou2025sweetrltrainingmultiturnllm}. 
Recent works have taken strides in developing diverse practical frameworks for multi-turn RL training of LLM agents \citep{wang2025ragenunderstandingselfevolutionllm, luo2025agentlightningtrainai, xi2025agentgymrltrainingllmagents, cao2025skyrlagentefficientrltraining, cui2025processreinforcementimplicitrewards, slime_github}. 
Across these systems, the training loop typically alternates between
(i) generating rollouts for a batch of prompts, i.e., trajectories of agent–environment interactions, and
(ii) updating the policy via policy gradient methods (e.g., PPO \citep{schulman2017proximalpolicyoptimizationalgorithms} and GRPO \citep{shao2024deepseekmathpushinglimitsmathematical}), using trajectory-level returns or per-turn rewards obtained from the environment.

Despite these advances, multi‑turn RL remains brittle as per-turn rewards can be sparse or delayed, environments may be stochastic, and interaction horizons are often long. 
Under these conditions, naive rollout sampling, i.e., independently sampling full-length trajectories from the current policy, can leave advantage estimates at the mercy of low‑quality or unlucky rollouts.
This can amplify variance, slow exploration, and precipitate mode collapse known as \emph{Echo Trap} \citep{wang2025ragenunderstandingselfevolutionllm}, where model performance abruptly declines during training due to spikes in gradients.
Several works have tried addressing the brittleness in RL, e.g., via turn-level credit assignment \citep{wei2025reinforcingmultiturnreasoningllm} and rejection-sampling-style rollouts such as GFPO \citep{shrivastava2025samplethinklessgroup}, RAFT \citep{dong2023raftrewardrankedfinetuning}, and StarPO \citep{wang2025ragenunderstandingselfevolutionllm}.
However, balancing exploration and exploitation, particularly in multi-turn environments, remains challenging.
We provide an overview of related works in \refapp{app:appendix_related_works}.

In this work, we ask the following fundamental question:
\emph{Given additional compute allocated specifically to rollout generation, how much can we improve the performance and stability of training if we restructure how trajectories are generated?}
Our answer is to move trajectory search from inference to training. 
Here, we draw inspiration from test-time scaling methods for reasoning \citep{snell2024scalingllmtesttimecompute}, which have shown that spending additional compute budget on search can reliably improve solution quality at inference time. 
Specifically, we ask whether a similar compute‑for‑quality trade‑off is applicable to multi-turn RL by upgrading the \emph{per-turn rollout generator} itself.

To this end, we introduce \textbf{TSR} (\textbf{T}rajectory‑\textbf{S}earch \textbf{R}ollouts), a train‑time framework that adapts test‑time search ideas to construct higher‑quality trajectories.
For each prompt, TSR runs a tree-style search over action sequences using available rewards or state-based feedback, and then retains the highest-scoring trajectories to compute advantages.
By optimizing the trajectory \emph{per-turn}, TSR avoids running into unlucky rollouts due to indiscriminate full-length trajectory sampling.
As a result, TSR changes only how rollouts are generated, making it optimizer-compatible by design.
This positions TSR within the family of search-guided policy improvement methods and is closely related in spirit to Expert Iteration~\citep{anthony2017thinkingfastslowdeep} and rejection-sampling-style fine-tuning, i.e., search improves the data distribution, while the policy optimizer distills the improved behavior back into the model.

In our experiments, we instantiate TSR with best-of-$N$, beam, and lookahead search methods, treating the corresponding search budgets (e.g., $N$, beam width, lookahead depth) as train-time compute knobs.
We evaluate TSR for Qwen2.5-0.5B and Qwen2.5-3B models \citep{qwen2025qwen25technicalreport} across three representative environments: (i) Sokoban \citep{weber2018imaginationaugmentedagentsdeepreinforcement}, a deterministic logic puzzle, (ii) FrozenLake \citep{brockman2016openaigym}, a stochastic RL game, and (iii) WebShop \citep{yao2023webshopscalablerealworldweb}, a long-horizon, agentic web navigation benchmark. 
TSR yields gains of up to 15 percentage points and converges in fewer optimization steps, without requiring additional search at inference time.

In summary, our main contributions are:
\begin{itemize}[leftmargin=*]
    \item \textbf{Search-Guided Rollout Generation.}
    We introduce TSR, an effective rollout generation framework for multi-turn RL that uses tree‑style search techniques inspired from test-time scaling to construct higher‑quality trajectories by optimizing rollouts at the \emph{per-turn level}.

    \item \textbf{Modular and Structured Design.}
    TSR changes only \emph{how} rollouts are generated, making it compatible with standard policy gradient methods (e.g., PPO, GRPO). We further show that its gains come from structured search rather than simply spending more compute on naive sampling.

    \item \textbf{Empirical Validation.}
    We demonstrate that TSR improves performance on deterministic, stochastic, and long-horizon tasks, enabling better performance with fewer interaction turns at inference.
\end{itemize}

%% file: chapters/2_background.tex
\section{Background and Problem Setup}
\label{sec:background}

We define preliminaries for multi-turn RL training and discuss the importance of high-quality rollouts.

\subsection{Multi-Turn RL as a Partially Observable Markov Decision Process (POMDP)}
Our focus is on agentic tasks, specifically multi-turn interactive decision-making, such as games, embodied navigation, or web-based interactions. 
Similar to \citep{xi2025agentgymrltrainingllmagents, luo2025agentlightningtrainai, zhou2024archertraininglanguagemodel}, we model such multi-turn decision-making problems as a \emph{Partially Observable Markov Decision Process} (POMDP) defined by the tuple $\mathcal{M} = \langle \mathcal{U}, \mathcal{S}, \mathcal{A}, \mathcal{O}, P, \mathcal{R}\rangle$, where $\mathcal{U}$ denotes the task space, $\mathcal{S}$ the state space, $\mathcal{A}$ the action space, and $\mathcal{O}$ the observation space.
$P:\mathcal{S}\times\mathcal{A}\rightarrow\mathcal{S}$ represents the state transition function and $\mathcal{R}:\mathcal{U}\times\mathcal{S}\times\mathcal{A}\rightarrow\mathbb{R}$ the reward function.

Given a task $u\in\mathcal{U}$, the LLM with policy $\pi_{\theta}$ parameterized by $\theta$ generates a first action $a_0\sim\pi_{\theta}(\cdot\mid u)$ and its state is transitioned according to $s_1\sim P(\cdot\mid s_0,a_0)$. 
A corresponding scalar reward $r_0 = \mathcal{R}(u,s_0,a_0)$ is provided to evaluate the quality of the action, and the agent receives an observation $o_0\in\mathcal{O}$ from the environment. 
At each time step $t$, given the history $\tau_{<t} = (a_0, o_0, r_0, \dots, a_{t-1}, o_{t-1}, r_{t-1})$, the agent generates the next action via $a_{t}\sim\pi_{\theta}(\cdot\mid u,\tau_{<t})$.
This process continues for a maximum turn-length (horizon) $K$, resulting in the trajectory $ \tau = (a_0, o_0, r_0, \dots , a_K, o_K, r_K) $ with corresponding cumulative reward $R(\tau) = \sum_{t=0}^{K} r_{t}$.

\subsection{Policy Optimization for Multi-Turn RL}
Policy gradient methods \citep{sutton1999policy} aim to maximize the expected cumulative reward, defined as the objective function $J(\theta) = \mathbb{E}_{\tau\sim\pi_{\theta}}\left[R(\tau)\right]$, via gradient ascent, i.e., $\theta_{\text{new}} = \theta + \eta \nabla_\theta J(\theta)$, where $\eta$ is the learning rate.
In the multi-turn setting, the policy gradient is given by $\nabla_\theta J(\theta) = \mathbb{E}_{\tau \sim \pi_\theta} \left[ \sum_{t=0}^{K-1} \nabla_\theta \log \pi_\theta(a_t \mid \tau_{<t}) \, \hat{A}_t \right]$, where $\hat{A}_t$ is the advantage estimate of taking action $a_t$.

\begin{wrapfigure}{r}{0.48\linewidth}
\vspace{-1em}
\centering
\begin{tcolorbox}[
    colback=gray!5, 
    colframe=magenta!70, 
    title=\textbf{Avoiding the ``Corner Trap" in Sokoban},
    boxrule=0.8pt,
    arc=2mm,
    fonttitle=\bfseries\small
]
\small

\textbf{Task:} Agent (A) must push box (B) into target (T). A ``Pillar'' blocks the path, requiring the agent to push the box downward first and then to the right to avoid a deadlock.

\vspace{0.2cm}

\noindent
\begin{minipage}{0.48\linewidth}
    \centering
    \footnotesize \textbf{A) Current State}\\[2pt]
    \begin{tikzpicture}[scale=0.6]
        \draw[step=1cm,gray!30,very thin] (0,0) grid (5,5);

        \fill[gray!70] (4,0) rectangle (5,5);
        \fill[gray!70] (0,4) rectangle (5,5);
        \fill[gray!70] (0,0) rectangle (5,1);
        \fill[gray!70] (0,0) rectangle (1,5);
        
        \fill[gray!70] (3,3) rectangle (4,4);
        \node[text=white, font=\tiny] at (3.5, 3.5) {Pillar};

        \node[text=red!70!black, scale=1.3] at (3.5, 1.5) {$\times$};
        \node[font=\tiny] at (3.8, 1.2) {\textbf{T}};

        \draw[fill=blue!30, thick] (2.1, 2.1) rectangle (2.9, 2.9);
        \node[font=\tiny] at (2.5, 2.5) {\textbf{B}};

        \draw[fill=red!30, thick] (1.5, 2.5) circle (0.35cm);
        \node[font=\tiny] at (1.5, 2.5) {\textbf{A}};

        \draw[->, ultra thick, red!80!black] (2.9, 2.5) -- (3.9, 2.5);
        \node[text=red!80!black, font=\tiny, fill=white, inner sep=2pt, opacity=1] at (2.2, 1.5) {``Bad Push"};

    \end{tikzpicture}
\end{minipage}
\hfill
\vrule width 0.5pt
\hfill
\begin{minipage}{0.48\linewidth}
    \centering
    \footnotesize \textbf{B) Deadlock State}\\[2pt]
    \begin{tikzpicture}[scale=0.6]
        \draw[step=1cm,gray!30,very thin] (0,0) grid (5,5);

        \fill[gray!70] (4,0) rectangle (5,5);
        \fill[gray!70] (0,4) rectangle (5,5);
        \fill[gray!70] (0,0) rectangle (5,1);
        \fill[gray!70] (0,0) rectangle (1,5);
        \fill[gray!70] (3,3) rectangle (4,4); 
        \node[text=white, font=\tiny] at (3.5, 3.5) {Pillar};

        \node[text=red!70!black, scale=1.3] at (3.5, 1.5) {$\times$};
        \node[font=\tiny] at (3.8, 1.2) {\textbf{T}};

        \draw[fill=blue!30, thick]  (3.1, 2.1) rectangle (3.9, 2.9);
        \node[text=red, font=\tiny] at (3.5, 2.5) {\textbf{B}};

        \draw[fill=red!30, thick] (2.5, 2.5) circle (0.35cm);
        \node[font=\tiny] at (2.5, 2.5) {\textbf{A}};

        \node[draw=red, text=red, font=\bfseries\scriptsize, fill=white, inner sep=2pt, rotate=0] at (2.1, 1.5) {STUCK};

    \end{tikzpicture}
\end{minipage}

\vspace{0.2cm}
\hrule height 0.5pt
\vspace{0.2cm}


\textbf{1. Naive Rollout.}
Agent prioritizes the immediate ``valid'' action: \emph{Push Right}.
\begin{itemize}[leftmargin=1.5em, noitemsep]
    \item \textbf{Mistake:} Box is pushed against the wall.
    \item \textbf{Deadlock:} To push the box down to ($\times$), the agent must stand on the cell above the box. However, that cell is occupied.
    \item \textbf{Outcome:} Box is frozen and the agent circles aimlessly until the episode ends. 
\end{itemize}

\textbf{2. Best-of-$4$ Rollout.}
Agent checks $N=4$ different possibilities (up, down, right, left).
\begin{itemize}[leftmargin=1.5em, noitemsep]
    \item \textbf{Exploration:} Some rollouts begin with \emph{Push Right} and reach the same deadlock, while others first reposition the agent.
    \item \textbf{Selection:} Only rollouts that preserve access above the box can complete the task.
    \item \textbf{Outcome:} Best-of-$4$ selects a feasible trajectory and avoids the irreversible trap.
\end{itemize}

\end{tcolorbox}
\caption{\textbf{``Corner Trap''.} Naive rollout sees \emph{Push Right} as progress but traps the box in a deadlock. Best-of-$N$ explores multiple possibilities and selects one that avoids the dead-end.}
\label{fig:sokoban_corner_trap}
\vspace{-6em}
\end{wrapfigure}

In general, directly computing policy gradients is difficult because the gradient estimator typically suffers from high variance.
To mitigate this, modern policy optimizers, such as PPO and GRPO, constrain the update step size and employ variance-reduction baselines for more reliable credit assignment. 
We provide details on how PPO and GRPO estimate advantages for multi-turn RL in \refapp{app:ppo_grpo_multi_turn_rl}.

In this work, we do not alter these algorithms but instead modify the \emph{rollout generation}.
Rather than passively sampling complete trajectories from the current policy $\pi_\theta$, we optimize rollouts on a per-turn level via search methods to improve the quality of the rollout data used for policy optimization.

\subsection{Quality, Diversity, and Stability}
The effectiveness and stability of multi-turn RL depend critically on the properties of the \emph{rollout distribution}, which directly controls the signal-to-noise ratio of advantage estimates, and which regions of the state-action space receive a learning signal \citep{schulman2015gae,schulman2017proximalpolicyoptimizationalgorithms,mroueh2025grpo}. 
In agentic environments, however, naive stochastic rollouts often suffer from several issues:

\vspace{-1em}

\begin{enumerate}[leftmargin=*]

    \item \textbf{Sparse or Delayed Rewards.}
    Per-turn rewards can be sparse, binary, or even delayed until the end of the episode.
    Consider a web agent that must purchase a specific item.
    Success requires a precise sequence of actions: \emph{search for item, click on item page, click on right color, click on buy}.
    If the agent executes the first three steps perfectly but fails to click ``Buy Now'' at the final step, the reward at the trajectory level becomes zero.
    This makes advantage estimation challenging as Monte Carlo returns can have high variance and bootstrapped estimators depend on value predictions that are hard to learn when intermediate rewards are uninformative \citep{schulman2015gae,schulman2017proximalpolicyoptimizationalgorithms}.

\end{enumerate}

\vspace{-2em}

\begin{enumerate}[leftmargin=*,resume]

    \item \textbf{Irreversible Traps.}
    In many multi-turn environments, early mistakes can create irreversible states in which the episode becomes unsalvageable, collapsing learning \citep{ignatenko2023sokoban_deadend}.
    To illustrate this, \refig{fig:sokoban_corner_trap} shows how a single early mistake in the Sokoban game can render an episode unsalvageable, whereas simple best-of-$N$ rollout selection recovers a feasible trajectory.

    \item \textbf{Insufficient Diversity.}
    On-policy methods can become unstable if the collected rollouts lack diversity.
    For example, if the agent repeatedly samples the same successful trajectory for a given state, the relative advantage becomes zero ($\hat{A}_t \approx 0$), effectively stalling gradient updates.
    This lack of exploration leads to brittle policies that overfit to specific actions or shortcut behaviors rather than learning robust, generalizable heuristics.
    Thus, effective training requires a rollout distribution that balances \emph{exploitation} (finding high rewards) with sufficient \emph{exploration} (covering diverse logic paths) to prevent the policy from getting stuck in local optima.

\end{enumerate}

These issues suggest that the \emph{quality of rollouts} is a central factor in multi-turn RL.
Motivated by the example in \refig{fig:sokoban_corner_trap}, we study whether allocating additional training-time compute to trajectory search can address these issues.
Specifically, we formulate the following hypothesis:

\begin{hypothesisbox}{Hypothesis: Search-Guided Rollout Optimization Improves Multi-Turn RL}
    In sparse-reward, multi-turn POMDPs, replacing naive stochastic rollout sampling with search-guided, per-turn trajectory optimization yields a rollout distribution with higher effective learning signal and improved stability, without modifying the policy optimizer.
\end{hypothesisbox}

We test this hypothesis by augmenting rollout generation via tree search guided by available step-level rewards or state-based feedback.
These signals are used only for rollout construction, whereas policy optimization remains based on the original task reward.
This repurposes test-time search for train-time data collection, analogous to recent process-reward-guided tree search methods for collecting higher-quality reasoning traces \citep{yao2023tot,zhang2024rest_mcts}.

%% file: chapters/3_trajectory_search_rollouts.tex
\section{Optimizing Rollouts with Trajectory Search}
\label{sec:trajectory_search}

In this section, we formalize rollout generation as a trajectory search problem and introduce TSR.

\subsection{Trajectory Rollouts via Tree Search}
\begin{figure*}
    \centering
    \includegraphics[width=0.8\linewidth]{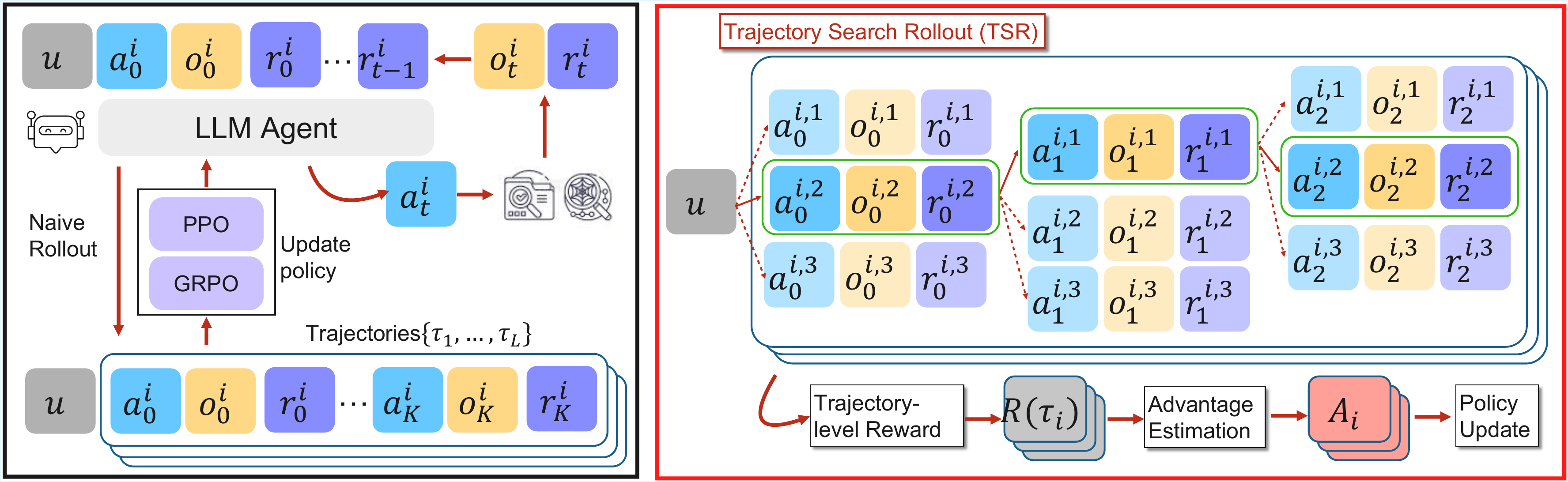}
    \caption{(\textbf{Left}) Multi-turn RL with naive rollouts: trajectories are sampled independently without any search. (\textbf{Right}) Trajectory Search Rollouts (TSR): using tree-style search to construct high-quality trajectories by  selecting high-scoring actions at each turn.}
    \label{fig:tsr_diagram}
    \vspace{-6pt}
\end{figure*}

We consider multi-turn rollout generation as a sequential decision problem that induces a search tree over action sequences (see \refig{fig:tsr_diagram}).
Starting from an initial state $s_0$, each node corresponds to a partial trajectory prefix $\tau_{<t}$, and edges correspond to candidate actions sampled from the current policy.

\paragraph{Candidate Action Set.}
At each turn $t$, instead of sampling a single action from the corresponding policy $\pi_\theta(\cdot \mid \tau_{<t})$, we sample a \emph{candidate action set} \( \mathcal{A}_t = \{ a_t^{(1)}, \dots, a_t^{(M)} \} \ , a_t^{(j)} \sim \pi_\theta(\cdot \mid \tau_{<t}) \), where $M$ controls the per-turn expansion budget.
This corresponds to expanding a node in the tree.

\paragraph{Scoring Function.}
We evaluate the desirability of taking action $a_t$ at prefix $\tau_{<t}$ with outcome $o_t$ from the environment using a \emph{scoring function} \( S(\tau_{<t}, a_t, o_t) \in \mathbb{R} \).
In agentic environments, $S(\cdot)$ scores the outcome of an executed action, e.g., the state, observation, or task progress, rather than arbitrary partial text prefixes.
When available, $S(\cdot)$ may simply correspond to the step-level environment reward.
When immediate rewards are sparse or delayed, it may instead incorporate heuristic signals, learned value estimates, or state-based feedback from the simulation.
This abstraction allows our framework to support
(1) \emph{reward-based scoring} in deterministic settings (e.g., logic puzzles such as Sokoban),
(2) \emph{heuristic or risk-aware scoring} in stochastic environments (e.g., FrozenLake), and
(3) \emph{progress-based or semantic scoring} in delayed-reward settings (e.g., WebShop).

\paragraph{Trajectory Search Rollouts (TSR).}
Given the candidate action set and scoring function, rollout generation selects actions according to a \emph{search strategy} $\mathcal{F}_{\phi}$ parameterized by search parameters $\phi$:
\begin{equation}
    \text{TSR} :
    (\pi_\theta, u, S, K, \mathcal{F}_{\phi})
    \;\longrightarrow\;
    \{\tau_1, \dots, \tau_L\} \ ,
\end{equation}
where $u\in\mathcal{U}$ denotes a task instance with initial state $s_0$ and $L$ denotes the number of retained trajectories.
By repeatedly expanding candidate actions and selecting according to the search strategy $\mathcal{F}_{\phi}$, which reweights the rollout distribution toward trajectories with higher-scoring prefixes, rollout generation corresponds to traversing a truncated tree.
Thus, TSR induces a search-guided rollout distribution $\mu_{\theta,\phi}(\tau \mid u; S)$.
Because all candidate actions are proposed by the current $\pi_\theta$, moderate search can remain policy-proximal (see \refsec{sec:distribution_shift} and \refapp{app:supplementary_results_distribution_shift}).
Crucially, TSR modifies only the \emph{rollout generation process}, i.e., the policy optimizer, objective, and update rules remain unchanged.

\subsection{Search Strategies}
We instantiate TSR with best-of-$N$, beam, and shallow lookahead search.
While these are popular test-time scaling techniques traditionally applied at inference time, here we adapt them to \emph{training-time rollout generation} for multi-turn RL.
\refig{fig:TSR_variants} in \refapp{app:TSR_search_strategies} provides an overview of these adaptations.

\paragraph{Trajectory-Level Best-of-$N$.}
For a given initial state, we independently sample a set of $N$ \emph{complete} trajectories
\(
\{\tau^{(1)}, \dots, \tau^{(N)}\}
\)
from the current policy $\pi_\theta$, where each trajectory is scored using its cumulative reward $R(\tau)$.
Rollout generation then retains the top-$L$ trajectories for policy optimization:
\begin{equation}
    \mathcal{T}^{\ast}
    =
    \operatorname{Top}_{L}
    \left(
        \{\tau^{(1)}, \dots, \tau^{(N)}\};
        R
    \right).
\end{equation}
Because selection occurs only \emph{after} full trajectories are generated, it does not influence intermediate action choices and may still generate suboptimal prefixes.
We therefore view trajectory-level best-of-$N$ as a flat search baseline with branching only at the root and selection after complete rollouts.

\paragraph{Per-Turn Beam Search.}
At each turn $t$, beam search maintains a set $\mathcal{B}_t$ of $B$ active partial trajectories and allocates a total budget of $M$ candidate actions across them.
Let $\mathcal{A}_t(\tau)$ denote the candidates sampled for prefix $\tau$, with $\sum_{\tau \in \mathcal{B}_t} |\mathcal{A}_t(\tau)| = M$.
Each candidate is scored using $S(\tau, a_t, o_t)$, and the top-$B$ resulting prefixes are retained:
\begin{equation}
    \mathcal{B}_{t+1}
    =
    \text{Top}_B
    \Big\{
    \tau \!\circ\! a_t
    \;\big|\;
    \tau \in \mathcal{B}_t,\;
    a_t \in \mathcal{A}_t(\tau)
    \Big\},
\end{equation}
where $\circ$ denotes concatenation and ranking uses accumulated scores.
After $K$ turns, the $B$ surviving trajectories are retained. 
To obtain $L$ trajectories per task group, we run $L/B$ independent beam searches.
Thus, beam search uses approximately $(L/B)MK$ candidate transitions compared with $LK$ for naive rollouts.
For $M=4$ and $B=2$, this yields a relative rollout-generation budget of $M/B=2\times$ over the standard instance-filtering baseline introduced in \refsec{sec:instance_filtering_for_enhanced_task_diversity}.

\paragraph{Shallow Lookahead Search.}
Similar to beam search, shallow lookahead evaluates candidate actions at each turn but additionally executes a short continuation of future steps (with depth $D \ll K$) before selection.
This corresponds to evaluating a truncated subtree rooted at $\tau_{<t}$ and choosing actions based on their downstream outcomes.
Because candidates and continuations are generated in a lightweight action-only format, its compute per update step remains comparable to beam search despite a larger transition count (see~\refapp{app:training_and_evaluation_setup}).

\subsection{Instance Filtering for Enhanced Task Diversity}
\label{sec:instance_filtering_for_enhanced_task_diversity}
While TSR improves the quality of trajectories (\emph{exploitation}), it does not by itself guarantee sufficient \textit{task-level} diversity (\emph{exploration}).
An established mechanism to promote such diversity is \emph{instance-level filtering} \citep{wang2025ragenunderstandingselfevolutionllm}, which hierarchically samples distinct task realizations before generating corresponding rollouts.
Specifically, at each training iteration, instance filtering first samples $P$ task instances (or \emph{groups}), where each group corresponds to a different initial state (e.g., different Sokoban layouts or WebShop tasks).
Within each group $u$ with initial state $s_0$, the agent generates $L$ corresponding rollout trajectories \mbox{$\{\tau^{(1)}, \dots, \tau^{(L)}\} \sim \pi_\theta(\cdot \mid u)$}, resulting in a total of $P \times L$ trajectories per iteration.
For each group, instance filtering then defines an \emph{outcome uncertainty} under the current policy $\pi_\theta$ as the standard deviation of resulting trajectory returns, i.e.,
\begin{equation}
    U(u; \pi_\theta) = \mathrm{Std}_{\tau \sim \pi_\theta(\cdot \mid u)} \big[ R(\tau) \big] \ .
\end{equation}

During training, groups are ranked according to $U(u; \pi_\theta)$, and only the top-$p\%$ most uncertain groups are retained for policy updates, while low-variance groups are discarded.
The motivation for group-based instance filtering stems from active learning \citep{settles2009active}, where groups with low reward variance tend to be uninformative, i.e., trajectories either solve trivial problems (uniformly high reward) or fail consistently (uniformly low reward), providing little learning signal.

\begin{wrapfigure}{r}{0.50\textwidth}
    \begin{minipage}{\linewidth}
    \vspace{-1em}
    \footnotesize
        \begin{algorithm}[H]
            \caption{TSR with Instance-Level Filtering}
            \label{alg:tsr_filtering}
            \begin{algorithmic}[1]
            
            \STATE \textbf{Input:} Policy $\pi_\theta$, Task distribution $\mathcal{D}$, Sample Size $P$, Trajectories $L$, Filter Ratio $p$, Strategy $\mathcal{F}_{\phi}$, Turn Horizon $K$, Score Function $S$.
            
            \STATE \textbf{Output:} Optimized Policy $\pi_{\theta^*}$
            
            \FOR{each training step}
            
                \STATE Sample $P$ task groups $\{u_1, \dots, u_P\} \sim \mathcal{D}$.
                
                \FOR{each group $u_i$}
                    \STATE \textcolor{neublue}{\emph{// Generate $L$ trajectories using TSR}}: \\
                    $\mathcal{G}_i \leftarrow\text{TSR}(\pi_\theta, u_i, S, K, \mathcal{F}_{\phi})$.
                    \STATE \textcolor{neublue}{\emph{// Compute outcome uncertainty}}: \\ 
                    $U_i = \mathrm{Std}(\{R(\tau) \mid \tau \in \mathcal{G}_i\})$.
                \ENDFOR
            
                \STATE Rank groups by $U_i$ and select top-$p\%$ $\mathcal{X}_{top}$.
            
                \STATE Update $\pi_\theta$ via PPO/GRPO using $\mathcal{X}_{top}$.
            
            \ENDFOR
            \end{algorithmic}
        \end{algorithm}
    \end{minipage}
    \vspace{-4.5em}
\end{wrapfigure}

Empirically, \citet{wang2025ragenunderstandingselfevolutionllm} show that uncertainty-based filtering can prevent mode collapse by maintaining a diverse training set.
We therefore combine TSR with instance filtering to jointly optimize rollout quality and diversity. 
\refalg{alg:tsr_filtering} summarizes the joint operation.

Since TSR is modular, it can be combined with other mechanisms that encourage exploration (e.g., ScalingInter-RL \citep{xi2025agentgymrltrainingllmagents}) or hierarchical RL methods (e.g., \citep{luo2025agentlightningtrainai}). 
A systematic study of integrating TSR with such methods is an interesting future work.

%% file: chapters/4_experimental_setup.tex
\section{Experimental Setup}
\label{sec:experimental_setup}

\subsection{Environments and Tasks}
\label{sec:environmets_and_tasks}
To ensure reproducibility, we base our experiments on RAGEN \citep{wang2025ragenunderstandingselfevolutionllm} and choose three diverse multi-turn environments to test agent performance under varying degrees of complexity:

\begin{enumerate}[leftmargin=*]
    \setlength{\itemsep}{0pt}
    \setlength{\leftmargin}{0pt}
    \item \textbf{Sokoban}.
    A logic puzzle where the agent must push boxes into target locations.
    Feedback is deterministic and instant, allowing to score per-turn rollouts via environment rewards. 

    \item \textbf{FrozenLake}.
    A stochastic navigation task where the agent must cross a frozen lake while avoiding obstacles (holes).
    The iced surface is slippery such that the same action can stochastically lead to different states.
    Rewards are extremely sparse, with a positive signal only upon reaching the goal.

    \item \textbf{WebShop}. 
    A deterministic e-commerce scenario in which the agent must navigate a website to purchase an item.
    Rewards are delayed as success is only signaled upon successful purchase, resulting in a long-horizon credit assignment problem.
    
\end{enumerate}

To address sparse or delayed rewards in FrozenLake and WebShop, we guide rollouts with light state-/progress-based scores, while optimizing only the original task reward.
We disentangle scoring from search in~\refapp{app:disentangling_search_from_scoring_signals} through reward-only, learned, task-agnostic, and information-matched controls.

\subsection{Training and Baselines}
We train Qwen2.5-0.5B and Qwen2.5-3B instruct models on Sokoban and FrozenLake, and only the larger Qwen2.5-3B model on WebShop due to the higher reasoning complexity required for this task.
We use PPO for Sokoban and GRPO for FrozenLake and WebShop, as GRPO has been shown to provide more stable learning for these tasks.
Following \citep{yu2025dapoopensourcellmreinforcement,wang2025ragenunderstandingselfevolutionllm}, we remove the KL term during optimization and apply asymmetric clipping (via clip-higher).
For each batch, we sample $P=16$ task groups (prompts) with a maximum turn horizon of $K=5$ for Sokoban/FrozenLake and $K=10$ for WebShop.
Our baseline is instance-level filtering with stochastic rollout sampling using a filtering ratio of $p=0.25$, similar to default RAGEN.
When instantiating TSR with best-of-$N$, we sample $N =32$ trajectories and retain the top $L=16$ trajectories per group.
For beam search, we use $M=4$ and beam width $B=2$.
For lookahead search, we use the same $M$ and $B$ with depth $D=2$. 
We selected these parameters after several ablations on $N$, $M$, $B$, and $D$ (see \refapp{app:supplementary_results}).
Additional details on training are provided in \refapp{app:training_and_evaluation_setup}.

\subsection{Evaluation and Metrics}
Across all three tasks, we report success rate as final task accuracy, evaluated on a held-out validation set of 256 fixed prompts per environment.
For inference, we use temperature $T = 0.5$ and truncate episodes after 5 turns for Sokoban and FrozenLake and 10 turns for WebShop.
Besides success rate (an indicator for \emph{task completion}), we also report rollout entropy (an indicator for diversity, i.e., \emph{exploration}) and average reward distribution for \emph{exploitation}. 
Training stability is analyzed via gradient norm statistics, where sudden spikes indicate potential mode collapse.
We additionally report average response length (\emph{reasoning verbosity}) and the average number of interaction turns (a proxy for \emph{interaction cost and task-completion latency}).
Further details on evaluations are given in \refapp{app:training_and_evaluation_setup}.

%% file: chapters/5_results.tex
\section{Results and Discussion}
\label{sec:results_and_discussion}

\begin{figure*}[t]
    \centering
    
    \begin{subfigure}[b]{0.32\linewidth}
        \centering
        \includegraphics[width=1\linewidth]{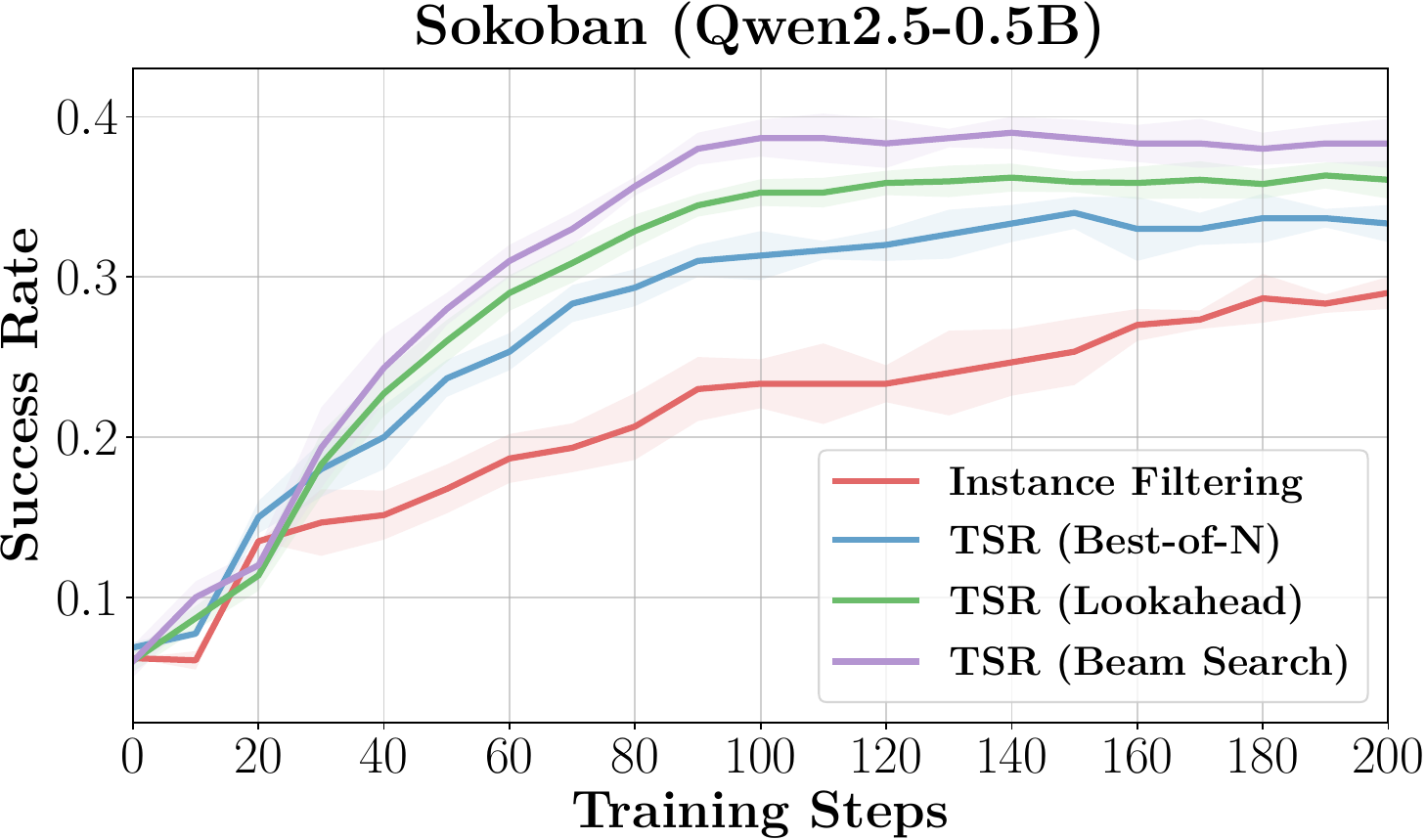}
        \caption{Sokoban (Qwen2.5-0.5B)}
    \end{subfigure}
    \hfill
    \begin{subfigure}[b]{0.32\linewidth}
        \centering
        \includegraphics[width=1\linewidth]{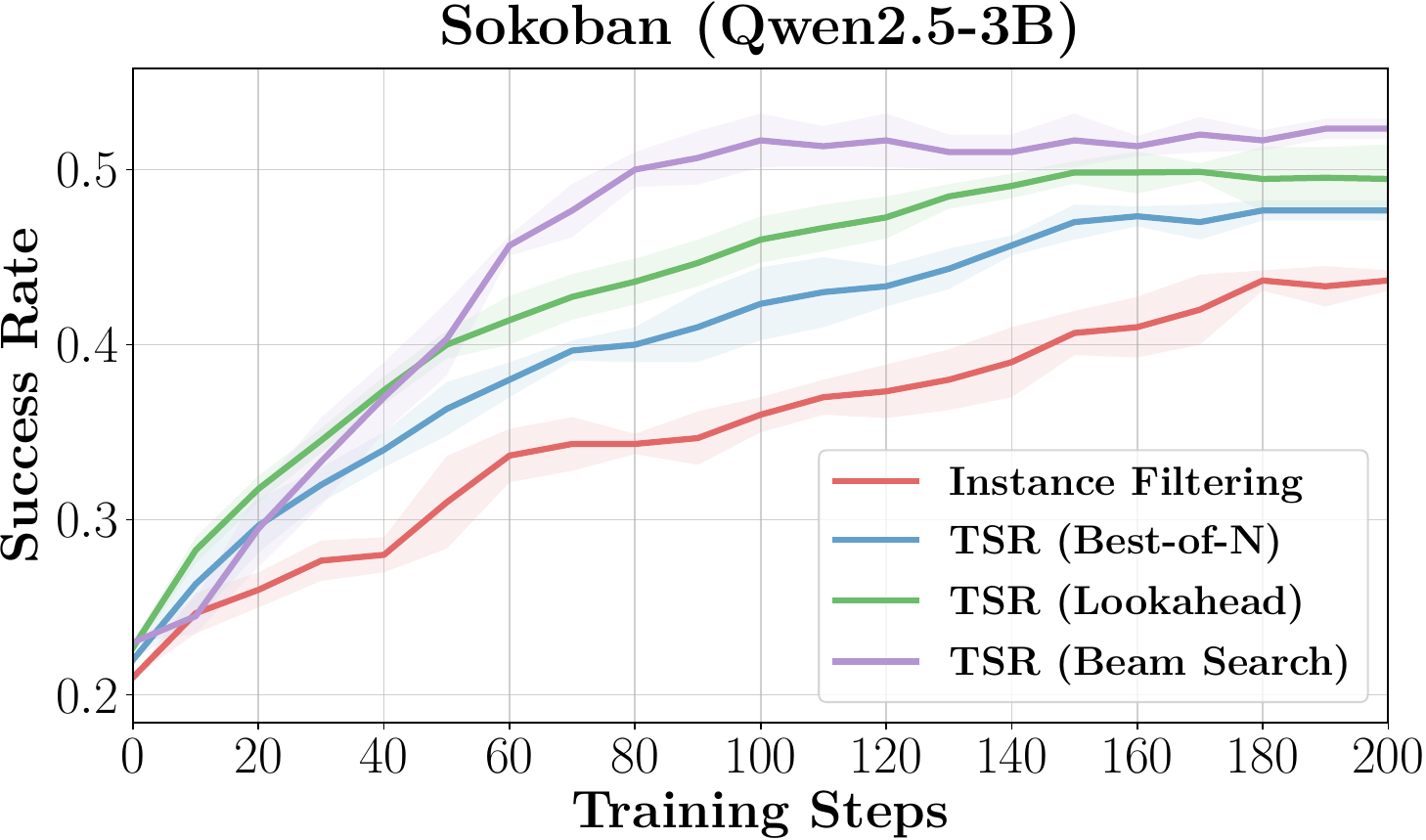}
        \caption{Sokoban (Qwen2.5-3B)}
    \end{subfigure}
    \hfill
    \begin{subfigure}[b]{0.32\linewidth}
        \centering
        \includegraphics[width=1\linewidth]{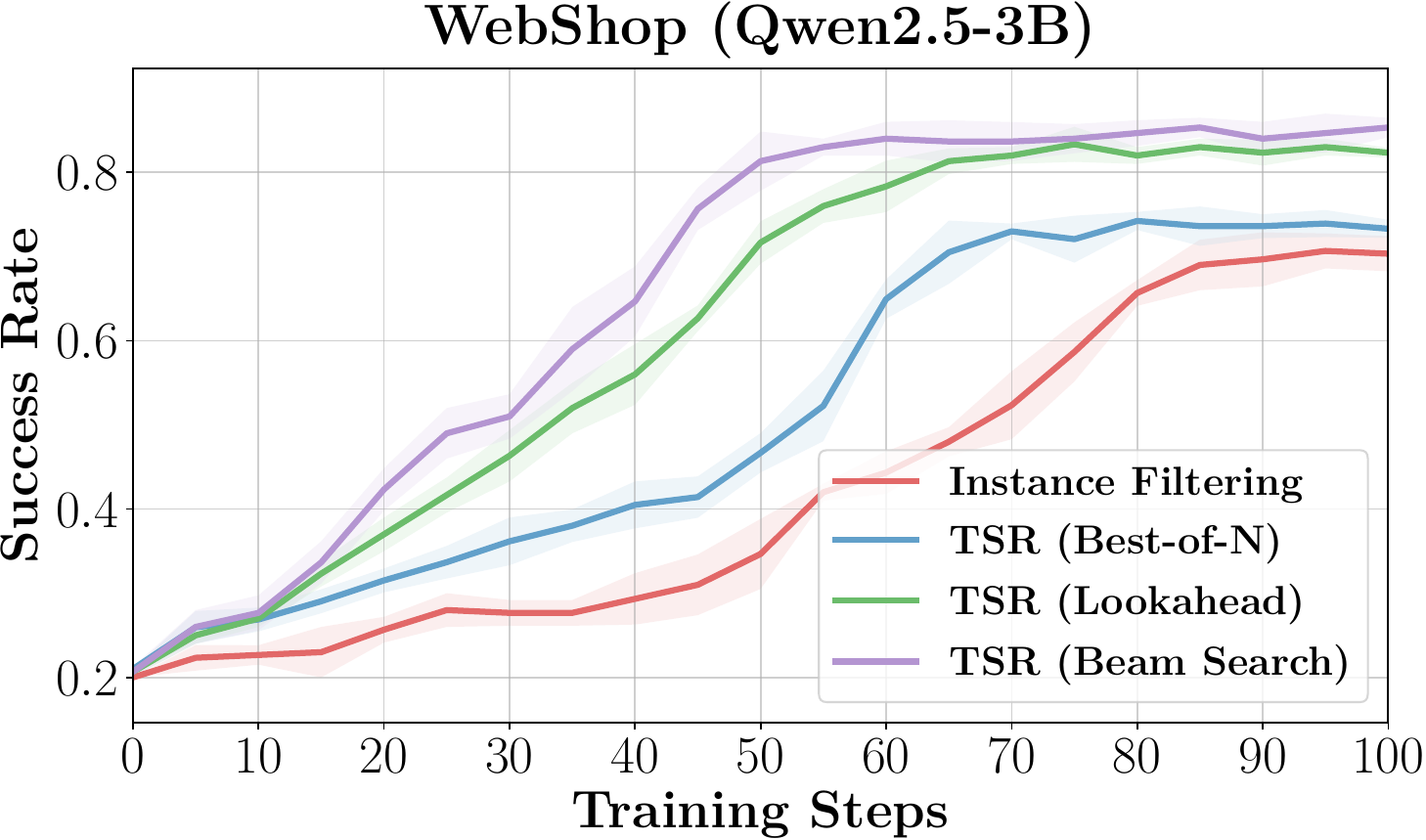}
        \caption{WebShop (Qwen2.5-3B)}
    \end{subfigure}
    
    \vspace{1em} 
    
    \begin{subfigure}[b]{0.32\linewidth}
        \centering
        \includegraphics[width=1\linewidth]{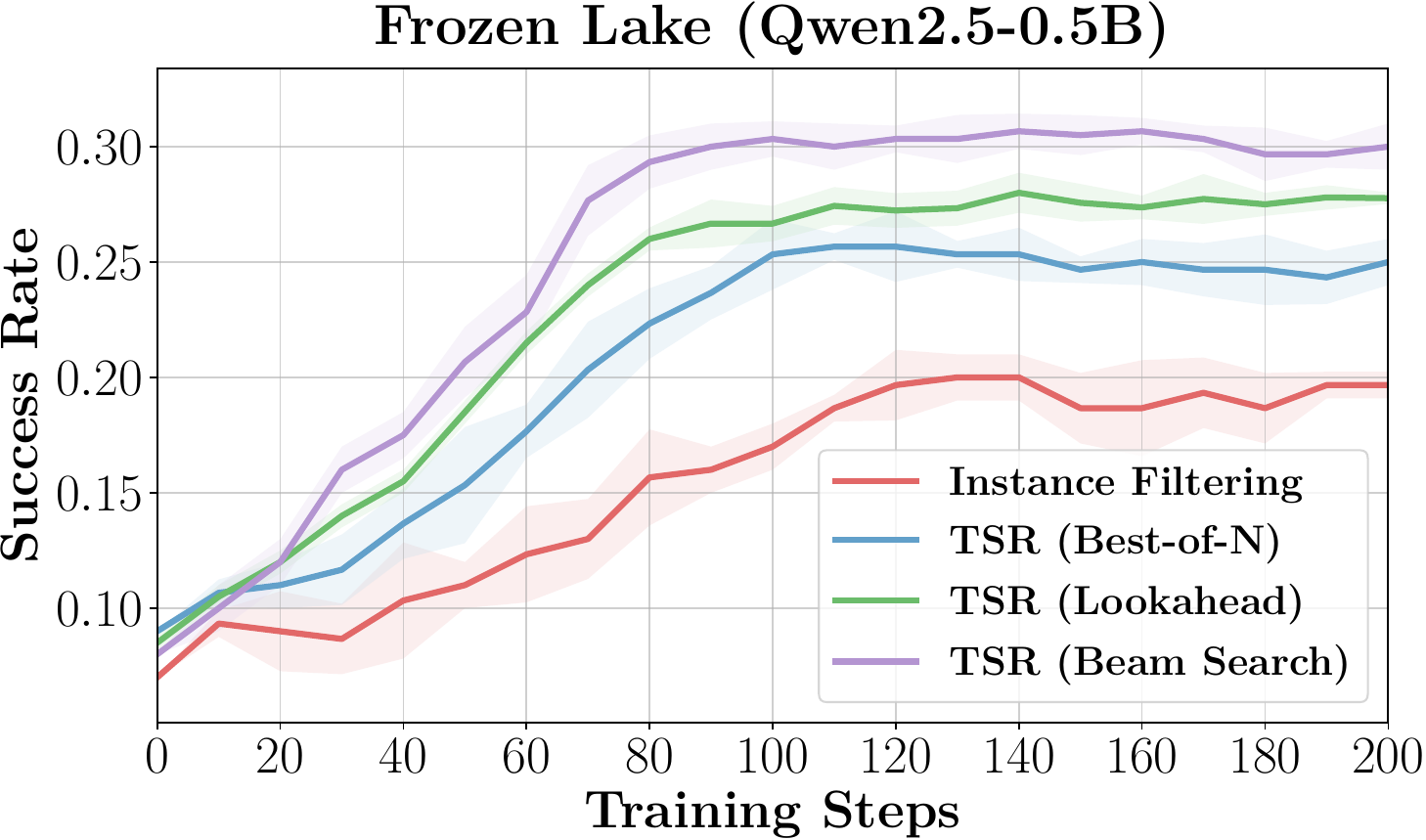}
        \caption{FrozenLake (Qwen2.5-0.5B)}
    \end{subfigure}
    \hspace{2em} 
    \begin{subfigure}[b]{0.32\linewidth}
        \centering
        \includegraphics[width=1\linewidth]{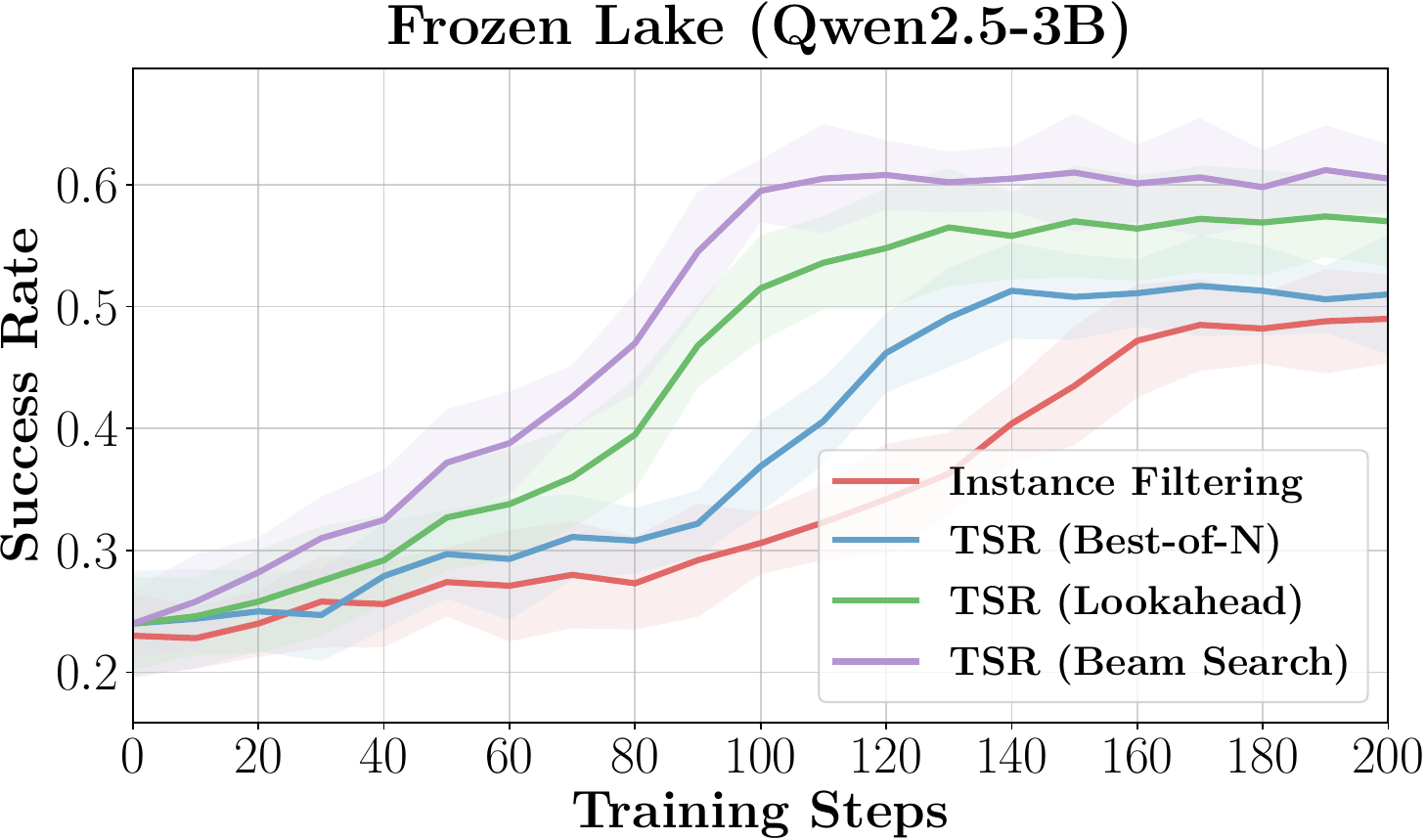}
        \caption{FrozenLake (Qwen2.5-3B)}
    \end{subfigure}

    \caption{\textbf{Success Rate Plots.} Comparison of TSR variants (Best-of-$N$, Lookahead, Beam Search) against the Instance Filtering baseline. Shaded regions show standard deviation across 3 runs.}
    \label{fig:success_rate_plots}
\end{figure*}

We present key results and statistics assessing final task performance, training stability, and scaling.
Further ablations, robustness analyses, and evaluations on other agentic tasks are provided in~\refapp{app:supplementary_results}.

\subsection{TSR Performance Comparison and Success Rates}
\begin{table*}[t]
\centering
\caption{\textbf{Results (Sokoban \& FrozenLake).} We report Success Rate ($\uparrow$; mean $\pm$ std), Average Response Length in tokens ($\downarrow$), and Average Interaction Turns ($\downarrow$) on the held-out validation set.}
\label{tab:main_results_sokoban_frozenlake}
\vspace{0.06in}
\small
\setlength{\tabcolsep}{3.5pt}
\renewcommand{\arraystretch}{1.1}

\resizebox{0.98\textwidth}{!}{%
\begin{tabular}{ll ccc c ccc}
\toprule
\multirow{2.5}{*}{\textbf{Task}} & \multirow{2.5}{*}{\textbf{Method}} & \multicolumn{3}{c}{\textbf{Qwen2.5-0.5B}} & & \multicolumn{3}{c}{\textbf{Qwen2.5-3B}} \\
\cmidrule{3-5} \cmidrule{7-9}
& & \small{Success Rate ($\uparrow$)} & \small{Resp. Len ($\downarrow$)} & \small{Turns ($\downarrow$)} & & \small{Success Rate ($\uparrow$)} & \small{Resp. Len ($\downarrow$)} & \small{Turns ($\downarrow$)} \\
\midrule

\multirow{5}{*}{\textbf{Sokoban}} 
& Base Model          
    & 8.9  & 293 & 4.7 
    & & 16.0 & 272 & 4.6 \\

& Instance Filtering  
    & $29.0 \pm 1.7$ & 105 & 4.4 
    & & $43.7 \pm 1.6$ & 161 & 4.1 \\

& TSR (Best-of-$N$)   
    & $33.3 \pm 1.5$ & 103 & 4.3 
    & & $47.7 \pm 1.5$ & 165 & 4.0 \\

& TSR (Lookahead)     
    & $36.1 \pm 1.4$ & 101 & 4.0 
    & & $49.5 \pm 1.4$ & 158 & 3.8 \\

\rowcolor{neugreenpastel!30} \cellcolor{white} 
& TSR (Beam Search)    
    & $\mathbf{38.3 \pm 1.3}$ & \textbf{98} & \textbf{3.8} 
    & & $\mathbf{52.3 \pm 1.3}$ & \textbf{152} & \textbf{3.6} \\

\midrule

\multirow{5}{*}{\textbf{FrozenLake}} 
& Base Model          
    & 6.4  & 276 & 4.6 
    & & 12.7 & 225 & 3.9 \\

& Instance Filtering  
    & $19.7 \pm 1.8$ & 157 & 4.1 
    & & $48.7 \pm 1.9$ & 174 & 3.6 \\

& TSR (Best-of-$N$)   
    & $25.0 \pm 1.6$ & 182 & 4.2 
    & & $51.0 \pm 1.6$ & 161 & 3.5 \\

& TSR (Lookahead)     
    & $27.8 \pm 1.5$ & 168 & 3.8 
    & & $57.0 \pm 1.4$ & 135 & 3.3 \\

\rowcolor{neugreenpastel!30} \cellcolor{white} 
& TSR (Beam Search)    
    & $\mathbf{30.0 \pm 1.4}$ & \textbf{152} & \textbf{3.5} 
    & & $\mathbf{60.7 \pm 1.2}$ & \textbf{98}  & \textbf{3.1} \\

\bottomrule
\end{tabular}
}
\end{table*}

We present success rate across training steps for all agent tasks in \refig{fig:success_rate_plots} and summarize overall task performance (final success rate) in Table~\ref{tab:main_results_sokoban_frozenlake} for Sokoban and FrozenLake, and in Table~\ref{tab:main_results_webshop} for WebShop.

Across all tasks and models, TSR methods consistently outperform the instance filtering baseline, which only promotes task-level diversity but does not optimize rollouts at the per-turn level.
For the 3B model, TSR (Beam Search) achieves a success rate of 52.3\% on Sokoban and 60.7\% on FrozenLake, improvements of 8.6 and 12 pp over instance filtering.
Similarly, TSR (Lookahead) and TSR (Best-of-$N$) achieve improvements ranging between 2--8 pp.
The largest gains are on WebShop, where best-of-$N$, lookahead, and beam search achieve performance improvements of 3, 12, and 15 pp, respectively, demonstrating that TSR yields strong benefits in this complex long-horizon setting.

Among TSR variants, beam search consistently achieves the strongest performance and converges faster (see \refig{fig:success_rate_plots}), followed by lookahead and best-of-$N$, which provide smaller gains.
Specifically, beam search maintains multiple partial trajectories, enabling recovery from early irreversible mistakes and providing robustness to environment stochasticity, particularly in FrozenLake and WebShop.

\begin{wraptable}{r}{0.52\textwidth}
    \vspace{-1em}
    \centering
    \caption{\textbf{Results (WebShop, Qwen2.5-3B).} Success Rate ($\uparrow$; mean $\pm$ std), Average Response Length ($\downarrow$), and Average Interaction Turns ($\downarrow$) on validation set.}
    \label{tab:main_results_webshop}
    \vspace{0.06in}
    \small
    \setlength{\tabcolsep}{4pt}
    \renewcommand{\arraystretch}{1.15}

    \resizebox{0.95\linewidth}{!}{%
    \begin{tabular}{l ccc}
    \toprule
    \multirow{2.5}{*}{\textbf{Method}} 
    & \multicolumn{3}{c}{\textbf{Qwen2.5-3B}} \\
    \cmidrule(lr){2-4}
    & \small{Success Rate ($\uparrow$)} 
    & \small{Resp. Len ($\downarrow$)} 
    & \small{Turns ($\downarrow$)} \\
    \midrule
    
    Base Model           
    & 3.0 & 747 & 7.7 \\
    
    Instance Filtering   
    & $70.3 \pm 1.8$ & 519 & 6.8 \\
    
    \midrule
    
    TSR (Best-of-$N$)    
    & $73.3 \pm 1.5$ & 504 & 6.5 \\
    
    TSR (Lookahead)      
    & $82.3 \pm 1.3$ & 475 & 6.1 \\
    
    \rowcolor{neugreenpastel!30} TSR (Beam Search)    
    & $\mathbf{85.3 \pm 1.1}$ & \textbf{453} & \textbf{5.8} \\
    
    \bottomrule
    \end{tabular}
    }
    \vspace{-1.5em}
\end{wraptable}

Lookahead commits to a single action after evaluating short-horizon continuations.
While effective in deterministic settings like Sokoban, its gains diminish as search depth increases due to higher cost of evaluating deeper subtrees.
Best-of-$N$ improves training by selecting higher-quality \emph{completed} trajectories, i.e., selection occurs only \emph{after} full rollouts are generated.
As a result, it acts as an oversampled variant of instance filtering, increasing the chance of ``luckier'' rollouts without actively steering intermediate decisions.

\subsection{Exploitation, Exploration, and Training Stability}
\begin{figure*}[t]
    \centering
    
    \begin{subfigure}[b]{0.32\linewidth}
        \centering
        \includegraphics[width=\linewidth]{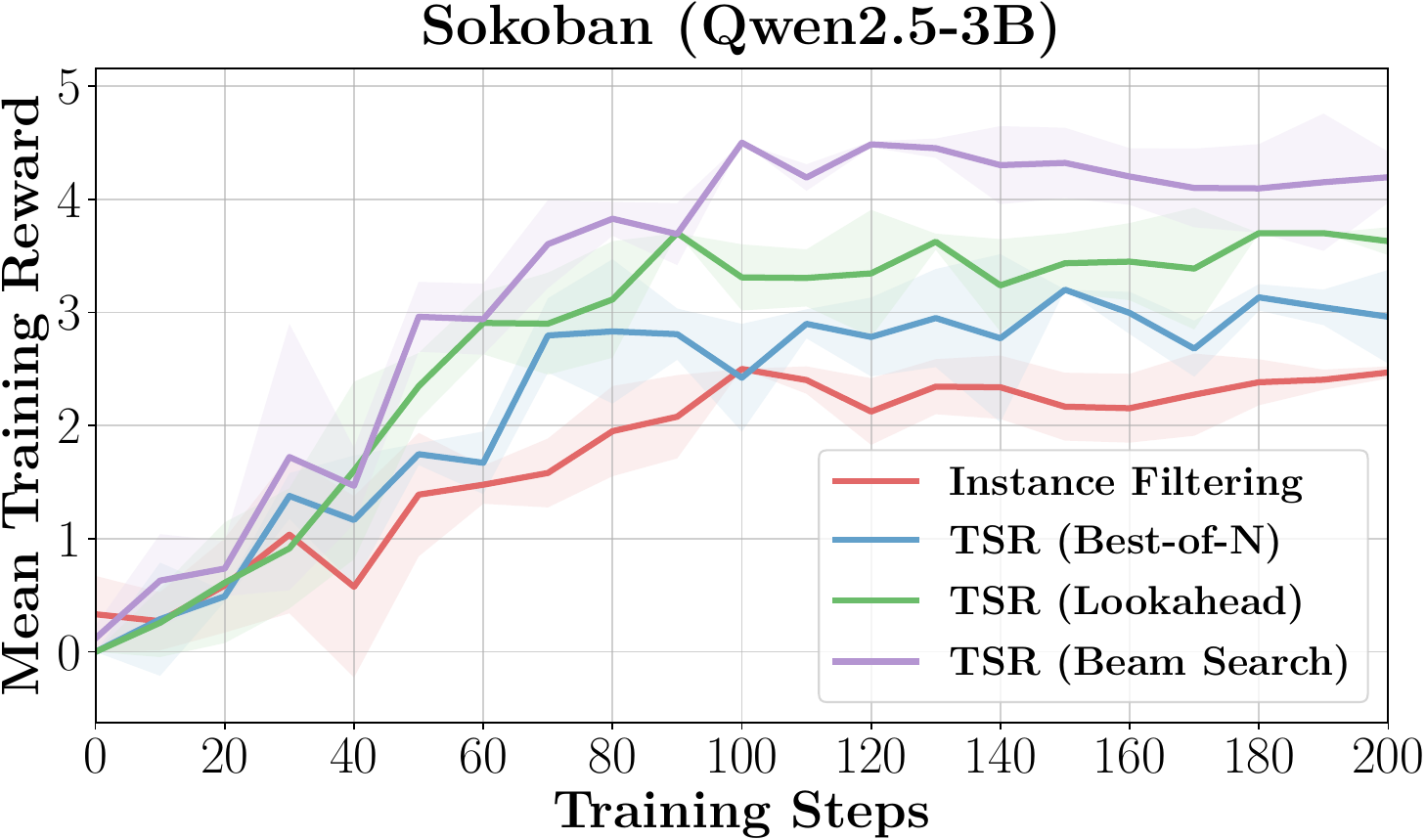}
        \caption{Average Reward (Exploitation)}
        \label{fig:average_reward}
    \end{subfigure}
    \hfill 
    \begin{subfigure}[b]{0.32\linewidth}
        \centering
        \includegraphics[width=\linewidth]{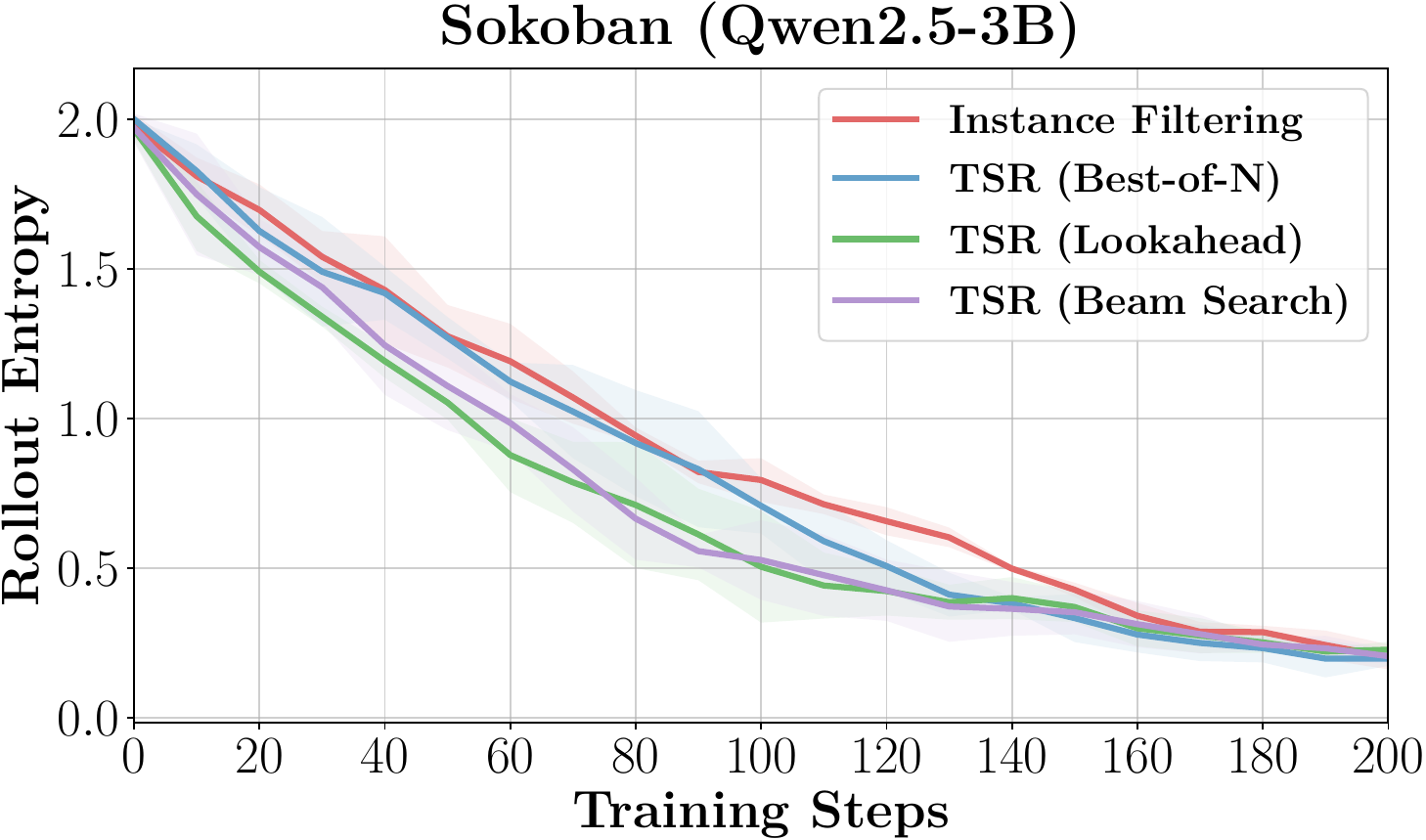}
        \caption{Rollout Entropy (Exploration)}
        \label{fig:rollout_entropy}
    \end{subfigure}
    \hfill
    \begin{subfigure}[b]{0.32\linewidth}
        \centering
        \includegraphics[width=\linewidth]{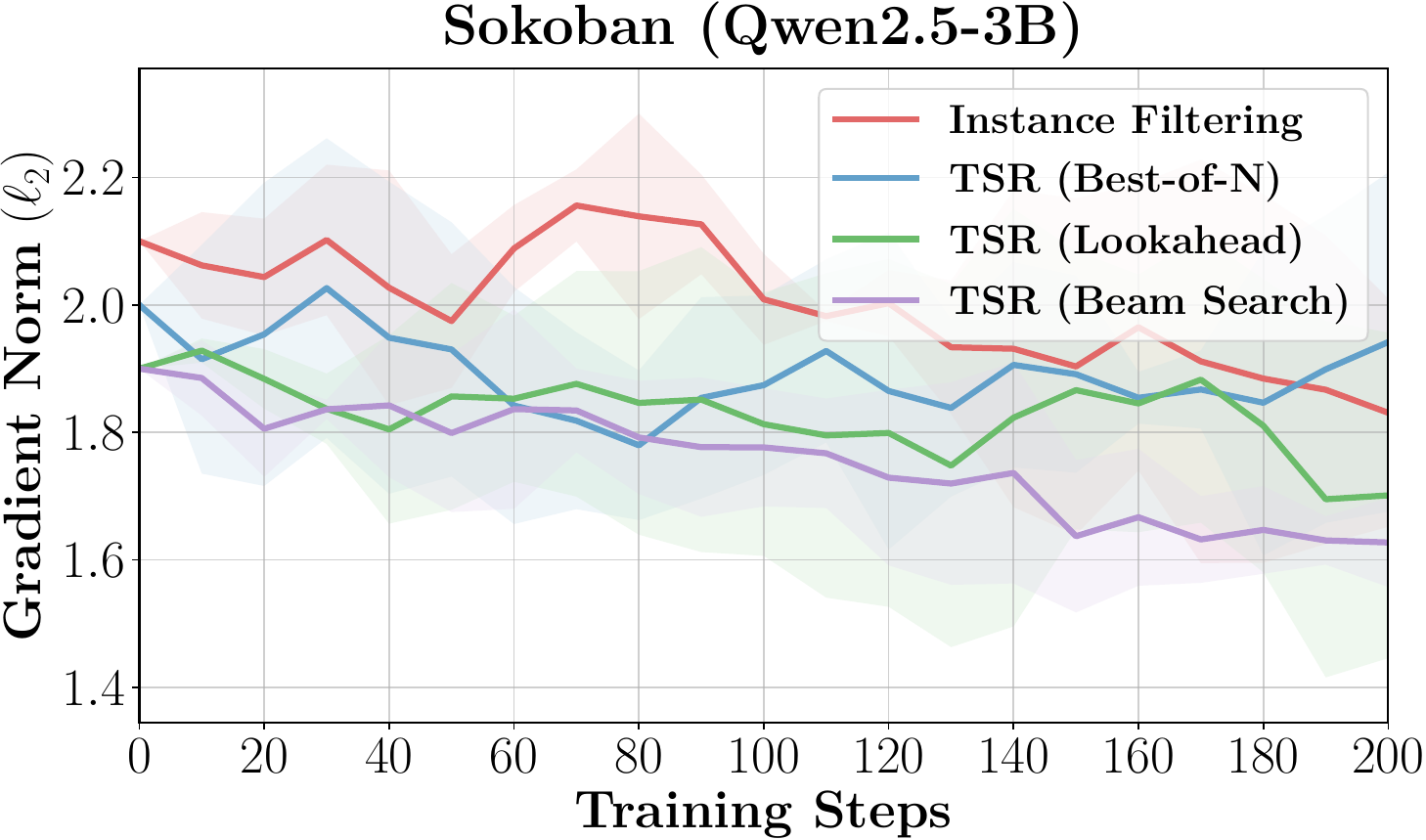}
        \caption{Gradient Norm (Stability)}
        \label{fig:gradient_norm}
    \end{subfigure}

    \caption{\textbf{Exploitation, Exploration, and Stability Metrics for Sokoban (Qwen2.5-3B).}
    \textbf{(a)} TSR achieves higher average rewards, indicating improved exploitation from higher-quality rollouts.
    \textbf{(b)} Rollout entropy decreases smoothly over training, suggesting sustained exploration, followed by policy consolidation.
    \textbf{(c)} Gradient norms remain stable and free of large spikes across TSR variants.}
    \label{fig:exploitation_exploration_stability_analysis}
    \vspace{-2em}
\end{figure*}

We analyze exploitation, exploration, and training stability using diagnostic statistics, shown exemplarily in \refig{fig:exploitation_exploration_stability_analysis} for the Sokoban task with the Qwen2.5-3B model.

For exploitation, we analyze average training rewards in \refig{fig:average_reward}.
Across all TSR variants, average rewards increase steadily over training, indicating effective learning and policy improvement.
The trends are further consistent with the overall task performance: beam search has on average higher rewards, followed by lookahead and best-of-$N$. 
This indicates that TSR shifts the exploitation bias toward higher rewards, resulting in stronger performance due to finding better trajectories.

For exploration, we analyze rollout entropy in \refig{fig:rollout_entropy}, which measures the uncertainty of the model's token predictions.
Across all methods, entropy decays smoothly over training, indicating that the agent is sufficiently exploring different strategies with uncertain outcomes in the beginning before converging to a more concentrated, exploitative behavior with a specific reasoning path.
Importantly, TSR variants exhibit a similar entropy decay profile to the instance filtering baseline, suggesting that search-guided rollout does not alter or destabilize the exploration–exploitation dynamics.

Finally, we analyze gradient norms in \refig{fig:gradient_norm} as a measure of stability. 
Sharp spikes in gradient norms are indicative of the ``Echo Trap'' and can lead to irreversible instability and mode collapse \citep{wang2025ragenunderstandingselfevolutionllm, xi2025agentgymrltrainingllmagents}. 
Across all methods, gradient norms remain without spikes.

\subsection{Inference Efficiency}
\label{sec:inference_efficiency}
In addition to final success rates, Tables~\ref{tab:main_results_sokoban_frozenlake} and \ref{tab:main_results_webshop} report average response length and interaction turns to assess whether train-time trajectory search  can improve solution quality, token efficiency, and interaction cost at inference time.
Across experiments, TSR consistently reduces both metrics, indicating improved decision-making at test time. 
In particular, beam search produces the shortest responses and fewest turns on average, with lookahead achieving comparable reductions.
Overall, this suggests that TSR distills more concise interaction strategies into the policy without requiring search at deployment.
We substantiate these results with qualitative examples for WebShop in \refapp{app:supplementary_results_inference_efficiency}.

\subsection{Rollout-Budget-Matched Baselines}
\label{sec:rollout_budget_matched_baselines}
Since TSR spends additional compute during rollout generation, we evaluate whether its gains can be matched by simply scaling naive rollout sampling.
Specifically, we keep the number of task groups $P=16$, the number of retained trajectories $L_{\mathrm{train}}=16$, the downstream optimizer, and the evaluation protocol fixed, while increasing the number of independently generated trajectories $L_{\mathrm{gen}}$.
For Qwen2.5-3B on FrozenLake, Table~\ref{tab:compute_scaled_if} shows that increasing $L_{\mathrm{gen}}$ from $16$ to $32$ and $64$, corresponding to $1\times$, $2\times$, and $4\times$ rollout-generation budgets, improves success only from $48.7\%$ to $49.5\%$ and $49.8\%$, respectively.
In contrast, TSR (Beam Search) reaches $60.7\%$ under the matched $2\times$ rollout-generation budget.
This substantiates that TSR's gains are not explained by additional sampling compute alone, but by allocating that compute structurally to search over intermediate decisions.
A more detailed rollout budget accounting is provided in \refapp{app:supplementary_results_compute_matched_baselines}.

\begin{table}[t]
    \centering
    \begin{minipage}[t]{0.48\textwidth}
        \vspace{0pt}
        \centering
        \caption{\textbf{Rollout-Budget-Matched Baselines (FrozenLake, Qwen2.5-3B).}
        Instance filtering plateaus, TSR uses compute more effectively.}
        \label{tab:compute_scaled_if}
        \resizebox{0.92\linewidth}{!}{%
        \begin{tabular}{lcc}
            \toprule
            \textbf{Method}
            & \textbf{Rel. Rollout Budget}
            & \textbf{Success Rate} \\
            \midrule
            Instance Filtering
            & $1.0{\times}$
            & $48.7 \pm 1.9$ \\

            Instance Filtering
            & $2.0{\times}$
            & $49.5 \pm 1.7$ \\

            Instance Filtering
            & $4.0{\times}$
            & $49.8 \pm 1.6$ \\

            \rowcolor{neugreenpastel!30}
            TSR (Beam Search)
            & $2.0{\times}$
            & $\mathbf{60.7 \pm 1.2}$ \\
            \bottomrule
        \end{tabular}
        }
    \end{minipage}
    \hfill
    \begin{minipage}[t]{0.48\textwidth}
        \vspace{0pt}
        \centering
        \caption{\textbf{Scaling Effects (Sokoban, Qwen2.5-3B).} Performance gains saturate as search budgets increase. $B=2$ offers the best trade-off.}
        \label{tab:scaling_summary}
        \resizebox{1\linewidth}{!}{%
        \begin{tabular}{llcc}
            \toprule
            \textbf{Method} & \textbf{Search Budget} & \textbf{Success Rate} & \textbf{$\Delta$} \\
            \midrule
            \multirow{3}{*}{\shortstack[l]{TSR\\(Beam Search)}} 
            & $M=2, B=1$  & $50.6 \pm 1.4$ & -- \\
            & \cellcolor{neugreenpastel!30}$M=4, B=2$  
            & \cellcolor{neugreenpastel!30}$\mathbf{52.3 \pm 1.3}$ 
            & \cellcolor{neugreenpastel!30}+1.7 \\
            & $M=6, B=2$  & $53.4 \pm 1.2$ & +1.1 \\
            \bottomrule
        \end{tabular}
        }
    \end{minipage}
    \vspace{-1em}
\end{table}

\begin{wraptable}{r}{0.45\textwidth}
    \vspace{-1em}
    \centering
    \caption{\textbf{Distribution Shift Analysis (FrozenLake, Qwen2.5-3B).} Moderate beam widths are policy-proximal, while excessive search increases KL, degrading performance.}
    \label{tab:kl_divergence}
    \resizebox{0.75\linewidth}{!}{%
    \begin{tabular}{ccc}
        \toprule
        \textbf{$B$} & \textbf{Avg. KL / Step} & \textbf{Success Rate} \\
        \midrule

        \rowcolor{neugreenpastel!30}
        $2$  & $\sim 0.02$ & $\mathbf{60.7 \pm 1.2}$ \\

        $5$  & $\sim 0.05$ & $61.3 \pm 1.1$ \\

        $10$ & $\sim 0.14$ & $60.9 \pm 1.4$ \\

        $20$ & $\sim 0.31$ & $53.8 \pm 1.8$ \\

        $30$ & $\sim 0.48$ & $45.1 \pm 2.1$ \\
        \bottomrule
    \end{tabular}
    }
    \vspace{-1.5em}
\end{wraptable}

\subsection{Search Budget, Performance, and Distribution Shift}
\label{sec:distribution_shift}
We further study how increasing train-time search budgets affects final performance.
As shown in Table~\ref{tab:scaling_summary}, we observe diminishing marginal returns.
Performance largely saturates as \(M\) increases, with differences between larger sampling budgets often within the observed run-to-run variation.
This justifies our choice of moderate budgets (e.g., $B=2$, $M=4$) for our experiments (see \refapp{app:supplementary_results_scaling_search_budgets}).

Increasing search budgets also affects the distribution shift between TSR's search-guided rollout distribution $\mu_{\theta,\phi}$ and the current policy $\pi_\theta$.
In general, PPO and GRPO accommodate this shift through clipping mechanisms, which enforce a trust region on each update.
To characterize when this shift becomes harmful, we perform an extreme beam width ablation and report the average KL divergence in Table~\ref{tab:kl_divergence}.
At moderate beam widths ($B \leq 5$), KL divergence remains low ($\lesssim 0.05$) and performance remains high.
At $B=10$, the distribution shift becomes more pronounced with only marginal gains, while wider beams degrade learning.
This shows that harmful distribution shift arises only far outside moderate beam widths.
Thus, in practical operating regimes, TSR behaves as a policy-proximal search procedure (see \refapp{app:supplementary_results_distribution_shift}).

%% file: chapters/6_conclusion.tex
\section{Conclusion}
\label{sec:conclusion}

In this work, we introduced Trajectory-Search Rollouts (TSR), a train-time rollout generation framework that adapts search methods, such as beam search, lookahead, and best-of-$N$, to enhance the training data quality in multi-turn agentic RL.
By exploring and pruning trajectories at the per-turn level, TSR changes only the rollout distribution and thus remains compatible with standard policy gradient optimizers.
Across Sokoban, FrozenLake, and WebShop tasks, TSR improves final task accuracy by up to 15 percentage points, while producing shorter responses and fewer interaction turns at inference time.
Our analysis further shows that 
(1) TSR improves the exploitation--exploration trade-off by steering rollouts toward higher-signal trajectories without destabilizing policy learning,
(2) TSR's gains stem from structured search rather than additional rollout budget alone, and
(3) TSR remains policy-proximal in practical operating regimes, while harmful distribution shift appears only under excessive search.
Thus, TSR demonstrates that moving search to train-time rollout construction can substantially improve performance, providing a modular way to trade additional train-time rollout compute for stronger deployment-time policies.
We discuss limitations and broader impact in \refapp{app:limitations}.

%% file: other/acknowledgements.tex
\section*{Acknowledgements}
This work was supported in part by the German Federal Ministry of Research, Technology and Space (BMFTR) within the research hub 6G-life (Grant 16KISK002) and through the project AISAC (Grant Number 16KIS2462), by the Bavarian Ministry of Science and the Arts and the Saxon Ministry for Science, Culture, and Tourism through the project Next Generation AI Computing (gAIn), by the Bavarian Ministry of Economic Affairs, Regional Development and Energy through the project 6G Future Lab Bavaria, and in part by IBM Research.

%% file: appendix/A_related_work.tex
\section{Related Works}
\label{app:appendix_related_works}

We present a short overview of relevant related works.

\paragraph{Multi-Turn Agentic RL Frameworks.}
Recent works have developed practical frameworks for training multi-turn LLM agents across diverse environments.
Agent Lightning~\citep{luo2025agentlightningtrainai} and AgentGym-RL~\citep{xi2025agentgymrltrainingllmagents} provide decoupled training systems and modular pipelines, building on AgentGym~\citep{xi2024agentgymevolvinglargelanguage}, which focuses primarily on environment diversity and data collection.
SkyRL~\citep{cao2025skyrlagentefficientrltraining} offers a full-stack library for tool-use agents with asynchronous weight updates to maximize throughput.
Similarly, slime~\citep{slime_github} decouples training from rollouts by connecting Megatron with SGLang, specifically targeting scaling of multi-turn RL.
RAGEN~\citep{wang2025ragenunderstandingselfevolutionllm} provides a trajectory-level RL formulation and documents instability patterns such as the Echo Trap, while their StarPO-S method mitigates collapse via instance filtering and gradient shaping.
TSR is complementary to these systems. 
It modifies the rollout generator through structured trajectory search while leaving the downstream optimizer, loss, and training framework unchanged.
We adopt instance filtering from RAGEN as a separate stabilizer in our experiments, while our main contribution is search-guided rollout generation within the training loop.

\paragraph{Test-Time Scaling and Rollout Search.}
Prior work has investigated test-time scaling strategies such as Best-of-$N$ sampling~\citep{ouyang2022traininglanguagemodelsfollow,gui2024bonbonalignmentlargelanguage}, beam search~\citep{wang2024offlinereinforcementlearningllm,alzorgan2025montecarlobeamsearch}, and shallow lookahead~\citep{lefebvre2025shallowplanningpartialobservability}.
These approaches allocate additional compute at inference time to improve solution quality.
TSR transfers this principle to training-time rollouts, where additional search is performed during training and distilled into a policy that requires no search at deployment.
Recent work has also extended test-time scaling to multi-turn agents by balancing search breadth against environmental-feedback depth~\citep{zhu2025scalingtesttimecomputellm} or scaling inference-time interaction and backtracking~\citep{shen2025thinkingvsdoingagents}.
Unlike these approaches, TSR places the additional search within rollout construction during RL training, allowing the resulting behavior to be internalized by the policy.

\paragraph{Search-Guided Policy Improvement and Expert Iteration.}
TSR is conceptually related to Expert Iteration~\citep{anthony2017thinkingfastslowdeep}, where tree search acts as a slow expert that improves upon the current apprentice policy and the neural network subsequently generalizes the improved plans back into a fast policy.
This search-and-distillation view also underlies AlphaZero-style~\citep{silver2017masteringchessshogiselfplay} policy improvement approaches.
TSR follows the same high-level principle for multi-turn LLM agentic RL, where search improves the rollout distribution during training and PPO/GRPO distill the resulting higher-quality behavior into the policy.
Unlike classical Expert Iteration, TSR does not rely on supervised imitation targets from a separate expert but instead uses search-guided trajectories directly within standard policy-gradient training.

\paragraph{Rejection-Sampling-Style Selection.}
Several training methods perform response- or trajectory-level filtering after samples have been generated.
For example, GFPO~\citep{shrivastava2025samplethinklessgroup} filters responses by properties such as length or reward-per-token within GRPO-style updates, while RAFT-style methods~\citep{dong2023raftrewardrankedfinetuning} train on positively rewarded samples.
TSR differs in that it targets interactive multi-turn agent RL and modifies trajectory construction before completion through per-turn branching, scoring, and pruning, rather than only filtering completed responses.
We view these approaches as complementary and compare TSR against search-free selection baselines under matched rollout-generation budgets.

\paragraph{Tree-Structured and Adaptive Rollouts.}
Several recent methods replace independent chain sampling with tree-structured rollouts, but they use the tree for different purposes.
Tree-GRPO~\citep{ji2026treesearchllmagent} expands randomly selected agent-step nodes to obtain more trajectories per token and tool-call budget, and derives intra- and inter-tree advantages from outcome rewards, so the tree serves sample efficiency and credit assignment.
ARPO~\citep{dong2025agenticreinforcedpolicyoptimization} and AEPO~\citep{dong2025agenticentropybalancedpolicyoptimization} branch at high-entropy tool-call steps to broaden exploration, coupled with modified advantage estimation and, for AEPO, a modified clipping rule.
LATR~\citep{xing2026lookaheadtreebasedrolloutsenhanced} branches at uncertain tokens and prunes near-duplicate continuations to increase diversity in single-turn RLVR.
In these methods, branching is driven by randomness or policy-intrinsic uncertainty, and pruning, where present, targets redundancy rather than return.
Concurrent work, PATR~\citep{li2026processrewardinformedtree}, scores partial trajectories every $K$ turns and gives promising branches more continuations, while all completed and pruned branches still enter the rollout group.
In all of these approaches, the tree shapes where rollout budget is spent but not which trajectories the policy learns from.
TSR instead treats rollout generation as a policy-improvement step.
Candidate actions proposed by $\pi_\theta$ are executed and scored at every turn, and search selects the trajectories that enter the update, so the policy learns from higher-return behavior than it would produce unaided.
Compared with PATR, TSR evaluates every decision point rather than periodic checkpoints, can score candidates by their downstream consequences through lookahead or Monte Carlo continuations, and instantiates a family of search strategies under a single rollout interface.
This applies the Expert Iteration principle inside on-policy agentic RL and raises a question that allocation- and exploration-oriented methods do not face, namely how far selection can shift the data away from $\pi_\theta$ before clipped updates stop benefiting.
While PATR notes that adaptive allocation biases the rollout group without quantifying this effect, we characterize the boundary explicitly in \refsec{sec:distribution_shift}.
Because TSR changes only which trajectories are trained on, it is complementary to these rollout schemes.

\paragraph{Turn-Level Credit Assignment.}
Recent work addresses sparse trajectory rewards by improving credit assignment within multi-turn trajectories.
MT-PPO/MT-GRPO~\citep{wei2025reinforcingmultiturnreasoningllm} introduce turn-level reward designs to provide denser supervision for multi-turn reasoning, while AgentPRM~\citep{xi2025agentprmprocessrewardmodels} uses process reward models to provide step-wise evaluations for long-horizon tasks.
GiGPO~\citep{feng2025groupingrouppolicyoptimizationllm} introduces a group-in-group advantage structure that combines episode-level relative advantages with step-level groups formed from repeated anchor states, enabling fine-grained credit assignment without additional rollouts.
TRACE~\citep{tao2026traceturnlevelrewardassignment} similarly targets long-horizon credit assignment by deriving dense turn-level rewards from changes in reference-model state values.
These methods primarily refine the learning signal or advantage estimator, whereas TSR focuses on improving the rollout distribution before the policy update.
We evaluate this distinction directly in~\refapp{app:comparison_to_related_RL_baselines}, where we compare TSR against GiGPO and additionally study their composition.

%% file: appendix/B_policy_optimization.tex
\section{Policy Gradient Methods in Multi-Turn RL}
\label{app:policy_optimization}

The core idea of policy gradient methods \citep{sutton1999policy} is to perform gradient ascent according to an objective $J(\theta)$, which typically represents the expected cumulative reward that the agent receives when following the policy $\pi_{\theta}$ and interacting with the environment.

\subsection{PPO and GRPO for Multi-Turn RL}
\label{app:ppo_grpo_multi_turn_rl}
We focus on two widely adopted approaches and adapt them to the multi-turn setting.

\paragraph{Proximal Policy Optimization (PPO).}
PPO \citep{schulman2017proximalpolicyoptimizationalgorithms} stabilizes training by limiting the magnitude of policy updates via a clipped surrogate objective:
\begin{align}
    J_{\text{PPO}}(\theta) = \mathbb{E}_{\tau \sim \pi_{\theta_{\text{old}}}} \Bigg[ \frac{1}{K} \sum_{t=0}^{K-1} \min \bigg( r_t(\theta) \hat{A}_t,
    \text{clip}(r_t(\theta), 1-\epsilon, 1+\epsilon) \hat{A}_t \bigg) \Bigg] \ ,
\end{align}
where
\(
r_t(\theta) = \frac{\pi_\theta(a_t \mid \tau_{<t})}{\pi_{\theta_{\text{old}}}(a_t \mid \tau_{<t})}
\)
is the turn-level importance-sampling ratio between the updated and behavior policy.
In standard PPO, $\hat{A}_t$ is typically computed using Generalized Advantage Estimation (GAE)~\citep{schulman2018highdimensionalcontinuouscontrolusing}, which relies on a learned value function (critic).

\paragraph{Group Relative Policy Optimization (GRPO).}
Learning a separate value function can still be computationally expensive and unstable.
GRPO~\citep{shao2024deepseekmathpushinglimitsmathematical} avoids an explicit critic by estimating advantages using relative comparisons.
Given a group of $G$ trajectories with returns $\{R_1, \dots, R_G\}$ sampled for the same task, the advantage for the $i$-th trajectory is computed as:
\begin{equation}
    \hat{A}_i
    =
    \frac{R_i - \mathrm{mean}(R_1, \dots, R_G)}
    {\mathrm{std}(R_1, \dots, R_G) + \epsilon} \ .
\end{equation}
GRPO then optimizes a PPO-style clipped objective averaged over the group:
\begin{align}
    J_{\text{GRPO}}(\theta)
    = \mathbb{E}_{\{\tau_i\}_{i=1}^G \sim \pi_{\theta_{\text{old}}}}
    \Bigg[
    \frac{1}{G}\sum_{i=1}^G
    \frac{1}{K_i}\sum_{t=0}^{K_i-1}
    \min\Big(
    r_{i,t}(\theta)\,\hat{A}_i,\, 
    \text{clip}(r_{i,t}(\theta), 1-\epsilon, 1+\epsilon)\,\hat{A}_i
    \Big)
    \Bigg] \, ,
\end{align}
where $K_i$ denotes the number of turns in $\tau_i$, and
\(
r_{i,t}(\theta)
=
\frac{\pi_\theta(a_{i,t}\mid \tau_{i,<t})}
{\pi_{\theta_{\text{old}}}(a_{i,t}\mid \tau_{i,<t})}
\)
is the per-turn importance-sampling ratio, defined analogously to PPO.

\subsection{Search-Guided Rollouts and Policy Gradient Compatibility}
\label{app:on_policy_analysis}
TSR modifies the data generation process used during rollout collection.
Instead of drawing complete trajectories directly from the current policy $\pi_\theta$, TSR samples candidate actions from $\pi_\theta$ and applies a search strategy $\mathcal{F}_\phi$ guided by a scoring function $S$.
This induces a search-guided rollout distribution
\begin{equation}
    \mu_{\theta,\phi}(\tau \mid u; S) \ ,
\end{equation}
which generally differs from the naive on-policy rollout distribution $\pi_\theta(\tau \mid u)$.
Thus, TSR should not be interpreted as an exact unbiased Monte Carlo estimator of the vanilla on-policy policy gradient objective.
Instead, TSR is best viewed as a policy-proximal, search-guided policy-improvement procedure.
This perspective is closely related to Expert Iteration and rejection-sampling-style fine-tuning, where a search or filtering procedure acts as an improvement operator over the current policy, producing higher-quality behavior than naive sampling.
The policy is then updated to distill this improved behavior.
Similarly, TSR uses $\pi_\theta$ to propose candidate actions, search to reweight trajectories toward higher-scoring prefixes, and PPO/GRPO to update the policy using the original task reward.
Importantly, TSR does not train directly on arbitrary off-policy data.
Instead, all candidate actions are sampled from $\pi_\theta$ and search only selects among these policy-generated candidates.

For GRPO-based experiments, this selection also changes the within-group return distribution used for group-relative advantage estimation.
This is most explicit for Best-of-$N$, where advantages are computed over the $L_{\mathrm{train}}$ retained trajectories rather than all $N$ generated candidates.
The selection therefore intentionally removes part of the low-return tail as part of the rollout-improvement operator.
Relative positive and negative advantages remain whenever the retained trajectories exhibit non-zero return variance, while groups with identical retained returns provide no group-relative learning signal and receive low uncertainty under the subsequent instance-filtering step.

TSR and instance filtering are therefore applied sequentially rather than independently.
For each task group, instance-filtering uncertainty is computed over the $L_{\mathrm{train}}$ trajectories returned by the rollout generator, i.e., over the post-search return distribution.
Search can consequently change within-group return variance and thereby affect which task groups are prioritized for policy updates.
We interpret instance filtering in this setting as measuring the residual outcome uncertainty after rollout improvement. 
Thus, groups whose post-search outcomes remain heterogeneous are prioritized, whereas groups that become uniformly successful or unsuccessful receive lower priority.
The same post-generation filtering rule is applied consistently across methods, although we do not separately disentangle this interaction in the present study.

Therefore, even though $\mu_{\theta,\phi}$ is not identical to $\pi_\theta$, the practical question is whether the induced shift remains small enough for clipped policy gradient updates to learn effectively from the resulting trajectories.
To this end, PPO and GRPO optimize clipped surrogate objectives whose probability ratios limit the effect of actions that become too unlikely under the updated policy.
As shown in \refsec{sec:distribution_shift} of the main body (see also \refapp{app:supplementary_results_distribution_shift}), in moderate search regimes, the selected trajectories remain policy-proximal while improving the quality of the rollout data.
However, if the search budget is made too large, $\mu_{\theta,\phi}$ can move too far from $\pi_\theta$, limiting the effective learning signal under clipping and reducing the usefulness of the rollout buffer.
This is precisely what we observe empirically: increasing the beam width initially improves performance while the average KL divergence remains small, whereas excessive beam widths substantially increase KL divergence and degrade performance.
These results support the view that TSR is effective when used as a moderate, policy-proximal search procedure rather than as an unconstrained search oracle.

%% file: appendix/C_search_strategies.tex
\section{Trajectory Search Rollout Search Strategies}
\label{app:TSR_search_strategies}

\begin{figure}[t]
    \centering
    \includegraphics[width=0.98\linewidth]{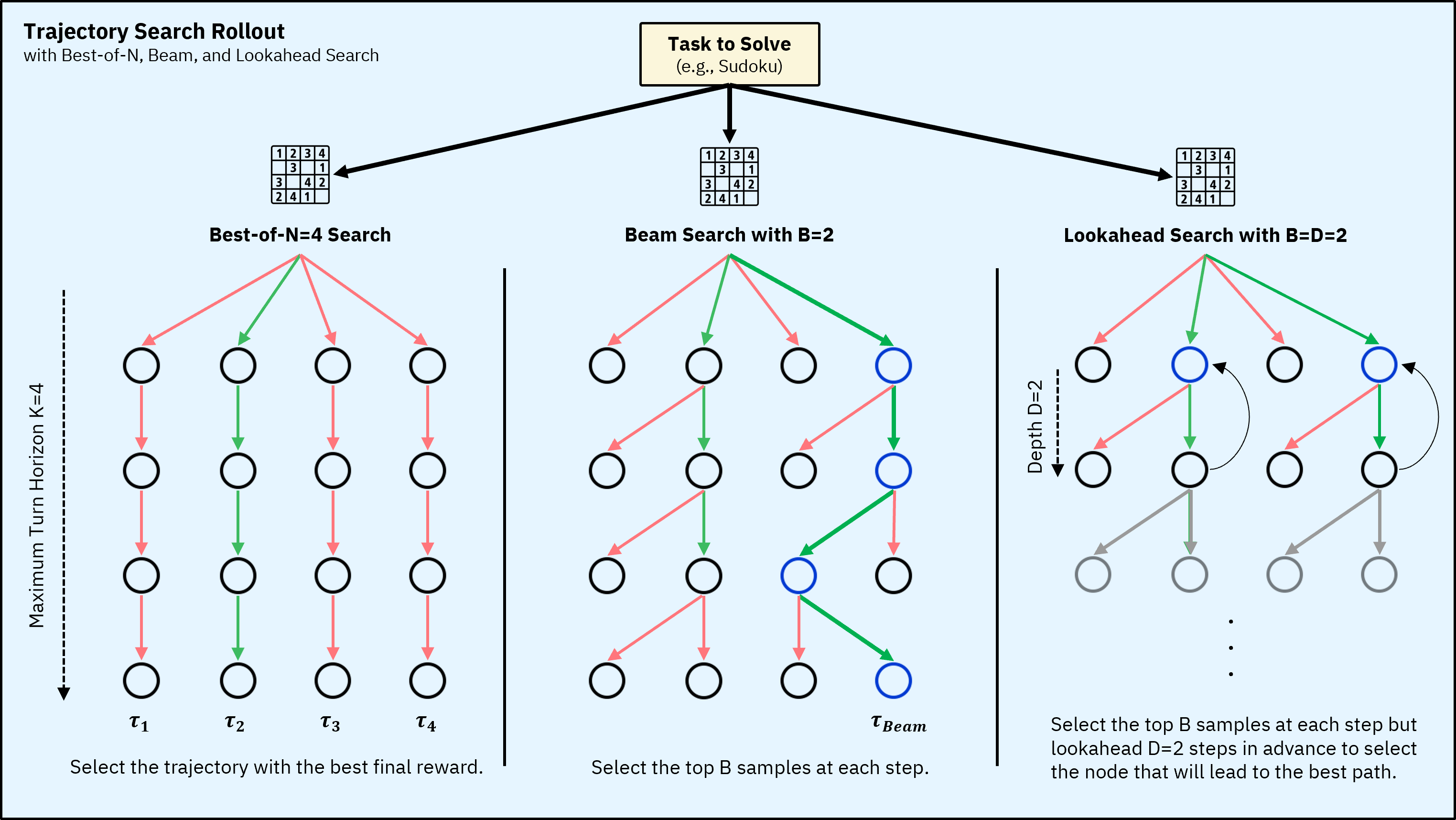}
    \caption{Illustration of TSR-adapted tree-search rollout generation for multi-turn RL training with best-of-$N$, beam search, and shallow lookahead search strategies.}
    \label{fig:TSR_variants}
\end{figure}

We instantiate TSR with three concrete search strategies inspired by test-time scaling~\citep{snell2024scalingllmtesttimecompute}: best-of-$N$, beam search, and shallow lookahead.
All three methods sample candidate actions from the current policy $\pi_\theta$ and use a scoring function $S$ to retain higher-scoring trajectories for policy optimization.
The examples below use Sokoban for explanation, where early box pushes can create irreversible deadlocks.
However, the same strategies apply generally to any multi-turn environment in which candidate actions can be evaluated using available rewards or state-based feedback.
\refig{fig:TSR_variants} provides a conceptual overview of how best-of-$N$, beam search, and shallow lookahead modify rollout generation in multi-turn RL.

\paragraph{Best-of-$N$.}
From a given initial state, the policy samples $N$ complete rollouts independently.
In Sokoban, some sampled trajectories may commit early to irreversible moves, such as pushing a box into a wall or corner, and therefore terminate with low return (see \refig{fig:sokoban_corner_trap}).
Other trajectories may preserve maneuverability and eventually solve the puzzle.
Best-of-$N$ scores the completed rollouts and retains the top-$L$ highest-scoring trajectories for policy optimization.
Because selection occurs only after full trajectories are generated, best-of-$N$ improves rollout quality but does not actively steer intermediate decisions.

\paragraph{Beam Search.}
Beam search performs selection at the level of partial trajectories.
At each turn, the policy samples candidate actions for each active prefix and TSR scores the resulting environment outcomes using $S$.
The search then retains the top-$B$ prefixes and prunes the rest.
In Sokoban, this allows beam search to discard prefixes that move toward dead-end configurations while preserving alternative branches that keep future pushes feasible.
Thus, beam search can recover from locally attractive but irreversible decisions by maintaining multiple candidate futures in parallel.

\paragraph{Shallow Lookahead.}
Shallow lookahead evaluates each candidate action by simulating a short continuation of depth $D \ll K$ before committing.
In Sokoban, an action whose immediate effect appears useful may quickly lead to blocked or low-mobility states after a few future pushes.
Lookahead avoids such actions when their short-horizon continuations score poorly, favoring actions that preserve access and future maneuverability.
Compared with beam search, shallow lookahead uses a shorter planning horizon and commits to a single action after evaluating the truncated subtree, reducing overhead while still improving over purely greedy-based rollout sampling.

%% file: appendix/D_environments.tex
\section{RL Environments}
\label{app:appendix_environments}

This section describes the Sokoban, FrozenLake, and WebShop environments used in our experiments.
We additionally describe the state-based scoring signals used by TSR for rollout construction in sparse- or delayed-reward settings.
Importantly, these signals are used only to guide search during rollout generation, i.e., policy optimization remains based on the original task reward.

\subsection{Environment Descriptions}
\label{app:environment_descriptions}

\begin{figure}[ht]
    \centering
    \begin{minipage}{0.32\textwidth}
        \centering
        \includegraphics[width=0.95\linewidth]{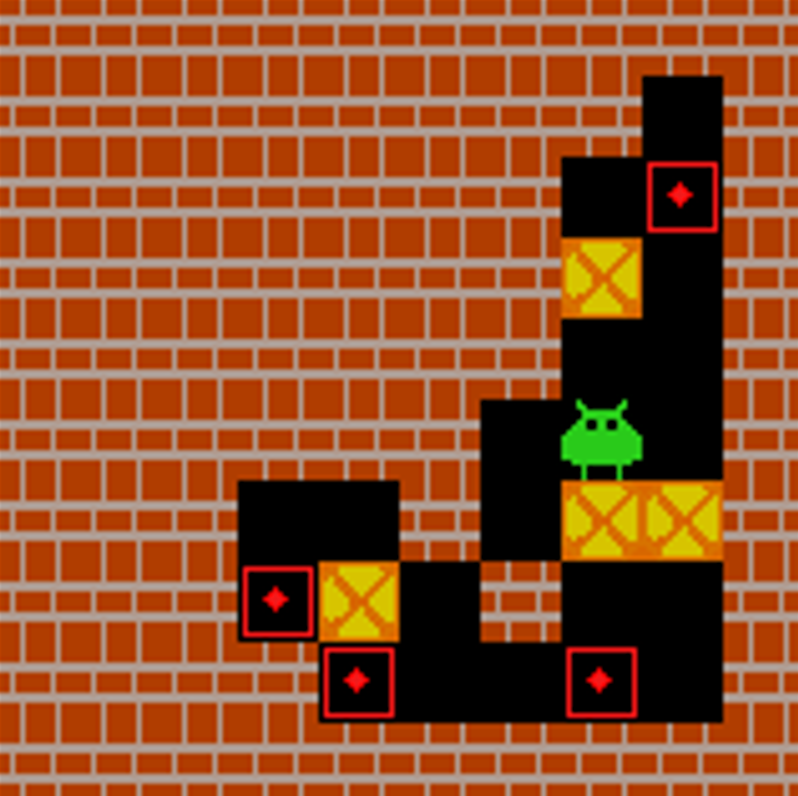}
        \caption{Sokoban.}
        \label{fig:sokoban-env}
    \end{minipage}
    \hfill
    \begin{minipage}{0.32\textwidth}
        \centering
        \includegraphics[width=0.95\linewidth]{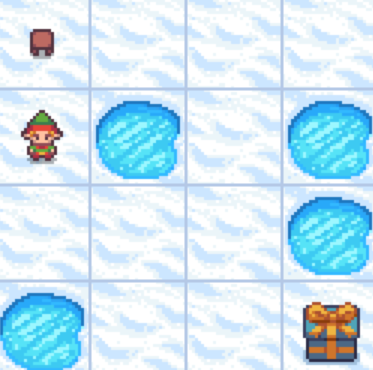}
        \caption{FrozenLake.}
        \label{fig:frozenlake-env}
    \end{minipage}
    \hfill
    \begin{minipage}{0.32\textwidth}
        \centering
        \includegraphics[width=0.95\linewidth]{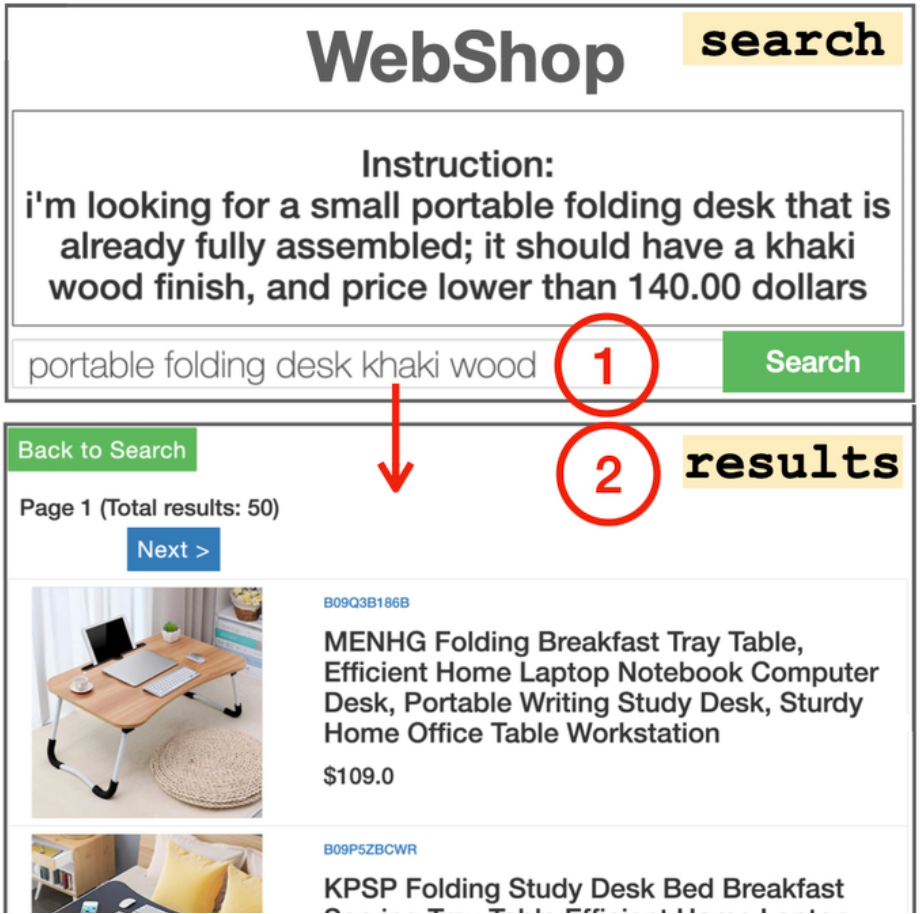}
        \caption{WebShop.}
        \label{fig:webshop-env}
    \end{minipage}
\end{figure}

\paragraph{Sokoban.} 
Sokoban~\citep{weber2018imaginationaugmentedagentsdeepreinforcement} is a deterministic logic puzzle in which the agent must push boxes onto designated target locations within a grid-based warehouse.
The agent navigates the grid and interacts with boxes via orthogonal pushes, but cannot pull boxes or push multiple boxes simultaneously.
The objective is to place all boxes on target cells while avoiding irreversible deadlocks, such as pushing boxes into corners or against walls where they can no longer be moved to a target.
In our setup (chosen from the RAGEN framework), rewards are provided at each step to capture both accuracy and efficiency: $+1$ per box on target, $-1$ per box off target, $+10$ on task completion, and $-0.1$ per action.
A sample environment is shown in \refig{fig:sokoban-env}.

\paragraph{FrozenLake.} 
FrozenLake~\citep{brockman2016openaigym} is a stochastic navigation task in which the agent must cross a grid-based frozen lake from a start state to a goal while avoiding holes.
The environment consists of safe frozen tiles, terminal hole states, and a terminal goal state.
The task is challenging because transitions are stochastic such that each action succeeds with probability $1/3$ and deviates perpendicularly with probability $2/3$ due to the slippery surface.
Rewards are sparse, with a positive reward of $+1$ only upon reaching the goal and $0$ otherwise.
A sample environment is shown in \refig{fig:frozenlake-env}.

\paragraph{WebShop.} 
WebShop~\citep{yao2023webshopscalablerealworldweb} is a deterministic e-commerce environment in which the agent must find and purchase a product satisfying a natural-language instruction.
The environment contains over one million real-world products with structured attributes, descriptions, and prices.
The agent interacts with the website through high-level actions such as searching, clicking product links, selecting options, and purchasing.
The task requires mapping the instruction's product constraints to available items while filtering irrelevant search results.
Rewards are delayed because success is determined only after the agent reaches the final purchase state with the correct item.
In our experiments, WebShop uses RAGEN's \texttt{webshop-minimal} adaptation.
A sample environment is shown in \refig{fig:webshop-env}.

\subsection{State-Based Scoring Signals for Rollout Construction}
\label{app:score_design}

\begin{table}[t]
    \centering
    \caption{\textbf{Scoring Signals Used by TSR.} State-based scores guide rollout construction only, whereas PPO/GRPO updates remain based on the original task reward.}
    \label{tab:score_signals}
    \resizebox{0.95\linewidth}{!}{%
    \begin{tabular}{lll}
        \toprule
        \textbf{Environment} & \textbf{TSR Rollout Scoring Signal} & \textbf{Policy Optimization Reward} \\
        \midrule
        Sokoban & Step-level environment reward / box-target progress & Original task reward \\
        FrozenLake & Goal-distance reduction and hole avoidance & Sparse goal-reaching reward \\
        WebShop & Milestone progress and keyword/attribute overlap & Final purchase success reward \\
        \bottomrule
    \end{tabular}
    }
\end{table}

In sparse- or delayed-reward environments, the environment reward may be uninformative for ranking candidate actions during intermediate rollout steps.
For example, FrozenLake provides a non-zero reward only upon reaching the goal and WebShop provides success feedback only after the final purchase.
To guide TSR in such settings, we use lightweight state-based scoring signals that evaluate the outcome of a candidate action after it is executed in the environment (see Table~\ref{tab:score_signals}).
These scores are not optimized as dense rewards by PPO/GRPO.
They are only used to rank candidate actions or trajectory prefixes during rollout construction, while policy optimization remains based on the original task reward.
Thus, TSR decouples \emph{search guidance} from \emph{policy optimization}: state-based scores help the rollout generator explore promising regions of the trajectory space, whereas the RL update continues to optimize the true task objective.
The resulting signals are deliberately simple and are derived from environment state variables already exposed by the simulator. 
More generally, scores may also be instantiated using other environment metadata, rule-based progress checks, learned value models, or process reward models.
Because these scoring signals provide additional training-time information, we explicitly isolate their contribution in~\refapp{app:disentangling_search_from_scoring_signals}.
Across reward-only, task-agnostic, learned, score-matched, and heuristic-free controls, TSR's structured search provides gains beyond access to the scoring signal itself.
In particular, Monte Carlo-backed beam search outperforms Best-of-$N$ using only the original terminal task reward, at additional rollout-generation cost, while an observation-only value model on WebShop shows that hidden simulator state is not required.
Thus, auxiliary scoring can strengthen TSR, but is not a prerequisite for effective trajectory search.

\paragraph{FrozenLake.}
FrozenLake has stochastic transition dynamics and provides a non-zero reward only upon reaching the goal.
Directly using the environment reward to rank intermediate candidate actions is therefore ineffective, since most non-terminal actions receive the same reward.
We instead use a simple state-based score that balances progress toward the goal with risk avoidance.
Let $s_t$ denote the agent's current grid position, $g$ the goal location, and $\mathcal{H}$ the set of hole states.
After executing action $a_t$ and observing the resulting state $s_{t+1}$, we define
\begin{equation}
\label{eq:frozenlake_score}
    S_{\text{FL}}(s_t, a_t)
    =
    - d(s_{t+1}, g)
    -
    \lambda \cdot \mathbb{I}[s_{t+1} \in \mathcal{H}] \ ,
\end{equation}
where $d(\cdot, \cdot)$ denotes the Manhattan distance, $\mathbb{I}[\cdot]$ is the indicator function, and $\lambda > 0$ controls the penalty for entering a hole.
This score favors actions whose realized next state is closer to the goal and penalizes unsafe transitions.
Because FrozenLake is stochastic, the score is applied to the observed post-transition state $s_{t+1}$, not to a deterministic prediction of the intended action outcome.

\paragraph{WebShop.}
WebShop is deterministic but exhibits delayed rewards since task success is only determined after the final purchase.
Consequently, the environment reward provides little guidance for ranking intermediate navigation actions.
We therefore use a lightweight state-based score based on task progress and instruction--page alignment.
Let $u$ denote the user instruction, $h_t$ the interaction history, and $p_t$ the textual content of the current webpage.
After executing action $a_t$, we define
\begin{equation}
\label{eq:webshop_score}
    S_{\text{WS}}(h_t, a_t)
    =
    \alpha \cdot \mathbb{I}[\text{progress}(h_{t+1})]
    +
    \beta \cdot \text{sim}(u, p_{t+1}) \ ,
\end{equation}
where $\text{progress}(\cdot)$ is a binary indicator for task milestones, such as reaching a product page, selecting required options, or entering checkout.
The term $\text{sim}(u, p_{t+1})$ is a low-cost textual similarity measure between the instruction and the resulting page content, implemented with simple keyword or attribute overlap.
The weights $\alpha,\beta>0$ balance structural progress and semantic alignment.
This score is intentionally simple: it is not intended to replace the task reward, but only to rank candidate branches during rollout generation.

%% file: appendix/E_training_setup.tex
\section{Training and Evaluation Setup}
\label{app:training_and_evaluation_setup}

Our training setup closely follows the experimental configuration of RAGEN~\citep{wang2025ragenunderstandingselfevolutionllm}. 
We summarize the key parameters and explicitly note deviations.
Unless stated otherwise, all environment settings, optimization details, and implementation choices are identical to those used in the default RAGEN configurations.

\paragraph{Models and Optimization.}
We initialize all agents from Qwen2.5-Instruct~\citep{qwen2025qwen25technicalreport} checkpoints.
For Sokoban and FrozenLake, we train both Qwen2.5-0.5B and Qwen2.5-3B models, while for WebShop we train only Qwen2.5-3B due to its longer context and higher reasoning demands.
Policy optimization is done using PPO for Sokoban and GRPO for FrozenLake and WebShop, consistent with \citet{wang2025ragenunderstandingselfevolutionllm}'s findings on stability under stochastic and sparse-reward settings.

\paragraph{Rollout and Update Configuration.}
Training proceeds for up to 200 rollout--update iterations for Sokoban and FrozenLake, and 100 iterations for WebShop.
At each iteration, we sample $P=16$ task instances (prompts) and retain $L_{\mathrm{train}}=16$ trajectories per task group before group-level instance filtering.
We distinguish these retained trajectories from additional candidates generated during rollout construction, denoted by $L_{\mathrm{gen}}$ when applicable.
For standard instance-filtered rollout sampling, $L_{\mathrm{gen}}=L_{\mathrm{train}}=16$.
Each trajectory is limited to a maximum of 5 interaction turns for Sokoban and FrozenLake and 10 turns for WebShop.
For TSR (Best-of-$N$), we sample $N=32$ complete candidate trajectories and retain the top $L_{\mathrm{train}}=16$, corresponding to a nominal complete-trajectory sampling budget of $N/L_{\mathrm{train}}=2\times$ relative to standard instance-filtered rollout sampling.
For TSR (Beam Search), we use a total per-turn expansion budget of $M=4$ across $B=2$ active beams.
The $M$ candidate expansions are not divided uniformly across the $B$ active beams but allocated according to their accumulated search scores, so higher-scoring prefixes can receive several candidates while lower-scoring prefixes receive none.
Consequently, even $M=B$ performs selection, and at $M=B=2$ beam search operates at a $1\times$ rollout-generation budget.
The $L_{\mathrm{train}}=16$ retained trajectories are obtained from $L_{\mathrm{train}}/B=8$ independent beam-search trees, each retaining $B=2$ terminal trajectories.
Beam search therefore executes approximately $(L_{\mathrm{train}}/B)MK$ candidate transitions per task group, compared with $L_{\mathrm{train}}K$ for standard instance-filtered rollout sampling, yielding a nominal relative rollout-generation budget of $M/B=2\times$.
For TSR (Lookahead), we run $L_{\mathrm{train}}=16$ independent searches, each producing one retained trajectory.
At every realized decision, we evaluate $M=4$ candidate actions and follow each candidate with one continuation of up to $D=2$ additional environment transitions.
The resulting nominal transition count per task group is
\[
T_{\mathrm{LA}}
=
L_{\mathrm{train}} M
\sum_{t=0}^{K-1}
\left[
1+\min(D,K-1-t)
\right],
\]
corresponding to approximately $9.6\times$ the naive transition count for Sokoban and FrozenLake ($K=5$) and $10.8\times$ for WebShop ($K=10$), before accounting for early termination.
To keep this tractable, candidate actions and continuations are generated without reasoning traces, while the retained trajectory keeps the lightweight outputs.
Consequently, although lookahead executes substantially more environment transitions, its measured compute is approximately $1.7$–$2.1\times$ that of instance-filtered sampling per update step (see Table~\ref{tab:convergence_analysis}).
Each root candidate is ranked by the undiscounted cumulative score along its continuation.
Throughout our compute accounting, all candidate generations and environment transitions used during search are counted, including branches that are evaluated but not retained for policy optimization.

\paragraph{Search and Scoring Hyperparameters.}
For FrozenLake, we set the hole-avoidance penalty in Eq.~\ref{eq:frozenlake_score} to $\lambda=5.0$, such that entering a hole is penalized more strongly than any single-step reduction in Manhattan distance.
For WebShop, we use $\alpha=1.0$ and $\beta=0.5$ in Eq.~\ref{eq:webshop_score}, giving greater weight to discrete task-progress milestones while using instruction--page similarity as a secondary ranking signal.
For the Monte Carlo-backed search in \refapp{app:heuristic_free_MC_search}, each candidate prefix is scored by the mean terminal task reward of its Monte Carlo continuations, which incurs additional rollout-generation compute beyond the matched $2\times$ comparisons.
For the noisy-score robustness experiments, we min--max normalize candidate scores to $[0,1]$ within each decision step before adding
$\epsilon\sim\mathcal{N}(0,\sigma^2)$.
Accordingly, $\sigma=0.25$ and $\sigma=0.50$ correspond to noise standard deviations of 25\% and 50\% of the normalized score range, respectively.

\paragraph{Rollout-Budget-Matched Baselines.}
To assess whether TSR's gains stem from structured search rather than raw rollout budget, we additionally evaluate compute-scaled instance-filtering baselines across all tasks with Qwen2.5-3B.
We keep the number of task groups $P=16$ and the downstream rollout set $L_{\mathrm{train}}=16$ fixed, while increasing the number of naively generated complete trajectories per task group from $L_{\mathrm{gen}}=16$ to $32$ and $64$.
These settings correspond to nominal rollout-generation budgets of $1\times$, $2\times$, and $4\times$, respectively, while leaving the downstream optimizer and evaluation protocol unchanged.
This isolates the effect of structured search from simply allocating additional generation compute to independent complete trajectories.
Actual wall-clock and total-compute costs are reported separately in \refapp{app:supplementary_results}.

\paragraph{Optimization Hyperparameters.}
We use the Adam optimizer with learning rate $1\times10^{-6}$ for both actor and critic (when applicable), and $(\beta_1,\beta_2)=(0.9,0.999)$.
For PPO, advantages are computed using Generalized Advantage Estimation (GAE)~\citep{schulman2018highdimensionalcontinuouscontrolusing} with $\gamma=1.0$ and $\lambda=1.0$, and a clipping parameter $\epsilon=0.2$.
For GRPO, advantages are computed via group-relative normalization over rollouts sampled from the same task instance.
Entropy regularization with coefficient $\beta=0.001$ is applied in all experiments.
Following RAGEN, we impose a response-format penalty of $-0.1$ when the model fails to produce a valid structured output.

\paragraph{Batching and Infrastructure.}
The update batch size is set to $E=32$, with a mini-batch size of 4 per GPU.
All experiments are conducted on 8 NVIDIA A100 GPUs using Fully Sharded Data Parallel (FSDP) training.
Rollout generation is accelerated using vLLM with retained computation graphs across prefill and sampling, and distributed execution is handled via Ray.

\paragraph{Evaluation Protocol.}
Evaluation is performed on a fixed set of 256 held-out prompts per environment.
Decoding uses temperature $T=0.5$, with episodes truncated after 5 turns for Sokoban and FrozenLake and 10 turns for WebShop.
We report the following metrics:

\begin{enumerate}[leftmargin=*]

    \item \textbf{Success Rate.}
    Final Task Accuracy on the fixed validation prompts, e.g., solving the puzzle in Sokoban, reaching the goal in FrozenLake, and completing a successful purchase in WebShop.

    \item \textbf{Rollout Entropy.}
    Average token-level entropy of sampled responses, used as an indicator for exploration and policy uncertainty.
    Sharp drops may indicate premature convergence or collapse.

    \item \textbf{Average Response Length.}
    Mean number of generated tokens per rollout, capturing verbosity and, in our setup, the extent of reasoning traces.

    \item \textbf{Gradient Norm.}
    $\ell_2$ norm of the policy gradient vector as an indicator of training stability.
    Sudden spikes correlate with unstable updates and potential collapse.

    \item \textbf{Average Number of Turns.}
    Average number of interaction turns per episode before termination (success, failure, or truncation), used to measure whether TSR reduces interaction budget at inference time after training.

\end{enumerate}

%% file: appendix/F_supplementary_results.tex
\section{Supplementary Results}
\label{app:supplementary_results}

In this section, we provide additional ablations and extended results.

\subsection{Inference Efficiency}
\label{app:supplementary_results_inference_efficiency}
Complementing our inference cost analysis in~\refsec{sec:inference_efficiency}, we provide an illustrative WebShop example showing how TSR can reduce redundant interactions at deployment time.
In particular, the instance filtering baseline enters a looping or backtracking pattern, whereas TSR (Beam Search) follows a shorter, more goal-directed trajectory.

\noindent\textbf{a) Instance Filtering: longer, looping behavior}
\begin{quote}\small
\textbf{Turn 1} \\
Instruction: \texttt{Find decorative pillows ... color beige latte ... size 20''x20'' ... under \$30 ...} \\
Output: \texttt{<think>Search for beige pillows with size 20x20.</think>
<answer>search[bisque pillow beige 20x20]</answer>} \\
Reward: \texttt{0.0}

\textbf{Turn 2} \\
State: Search results page (50 items). \\
Output: \texttt{<think>Click a result and then filter color/size.</think>
<answer>click[b08l2zdwn2]</answer>} \\
Reward: \texttt{0.0}

\textbf{Turn 3} \\
State: Filters shown (\texttt{color: beige, black, ...}, \texttt{size: 18''x18'', ...}). \\
Output: \texttt{<think>Try clicking color filters.</think>
<answer>click[beige latte]</answer>} \\
Reward: \texttt{0.0}

\textbf{Turn 4--8} \\
Output: \texttt{(repeated filter clicks / back-and-forth navigation)} \\
Reward: \texttt{0.0} \\
(\emph{Agent exhausts actions or remains in an unproductive loop.})
\end{quote}

\noindent\textbf{b) TSR (Beam Search): shorter, goal-directed behavior}
\begin{quote}\small
\textbf{Turn 1} \\
Output: \texttt{<think>Use a targeted query to reduce backtracking.</think>
<answer>search[decorative pillow cover beige latte]</answer>} \\
Reward: \texttt{0.0}

\textbf{Turn 2} \\
State: Search results page. \\
Output: \texttt{<think>Select the most compatible result, then finish checkout.</think>
<answer>click[b09f6scw47]</answer>} \\
Reward: \texttt{0.0}

\textbf{Turn 3} \\
State: Product page with options + \texttt{buy now}. \\
Output: \texttt{<answer>click[buy now]</answer>} \\
Reward: \texttt{+1.0}
\end{quote}

Similar qualitative patterns can be observed in other WebShop instances, with TSR exhibiting less redundant navigation and shorter interaction trajectories.

\paragraph{Interpretation of Inference-Efficiency Metrics.}
Average response length and interaction turns are measured over complete evaluation episodes and therefore are not independent of task success, since successful episodes may terminate earlier.
We consequently interpret these metrics as measures of the realized end-to-end inference cost of the learned policy, rather than as success-conditioned measures of efficiency.
Thus, the observed reductions capture the combined effect of improved task completion and more concise interaction behavior at deployment time.

\subsection{Rollout-Budget-Matched Baselines}
\label{app:supplementary_results_compute_matched_baselines}
Because TSR incurs additional compute during rollout generation, an important question is whether its gains can be matched by allocating comparable rollout-generation compute to naive stochastic sampling.
To this end, we extend our analysis in~\refsec{sec:rollout_budget_matched_baselines} and construct compute-scaled instance-filtering baselines across Sokoban, FrozenLake, and WebShop with Qwen2.5-3B.
For all comparisons, we keep the number of task groups $P=16$, the number of trajectories retained for policy optimization $L_{\mathrm{train}}=16$, the downstream optimizer, training schedule, evaluation protocol, and model architecture fixed.
For instance filtering, we increase the number of independently generated complete trajectories per task group from $L_{\mathrm{gen}}=16$ to $32$ and $64$, corresponding to nominal rollout-generation budgets of $1\times$, $2\times$, and $4\times$, respectively.
TSR (Beam Search) uses $M=4$ total candidate expansions per turn with $B=2$ active beams, corresponding to a nominal rollout-generation budget of $M/B=2\times$ relative to standard instance-filtered rollout sampling.

As shown in Table~\ref{tab:app_compute_scaled_if_all_tasks}, increasing the naive rollout budget yields only modest improvements across all tasks.
On Sokoban, increasing $L_{\mathrm{gen}}$ from $16$ to $32$ and $64$ improves success from $43.7\%$ to $44.8\%$ and $45.2\%$, whereas TSR (Beam Search) reaches $52.3\%$ under the matched $2\times$ rollout-generation budget.
On FrozenLake, the same scaling improves success only from $48.7\%$ to $49.5\%$ and $49.8\%$, while TSR reaches $60.7\%$ at $2\times$.
On WebShop, increasing the naive rollout budget improves success from $70.3\%$ to $72.1\%$ and $72.8\%$, while TSR reaches $85.3\%$ at $2\times$.
These results show that TSR's gains cannot be explained by additional rollout-generation compute alone.
Instead, structured search allocates this compute to intermediate decision points, allowing the rollout generator to avoid locally poor prefixes and construct higher-signal trajectories.
In contrast, naive scaling spends the additional compute on independently sampled complete trajectories and therefore cannot redirect rollout generation before irreversible or low-value decisions have already been made.
Overall, under the same downstream training setup, structured per-turn search makes more effective use of a matched rollout-generation budget than simply increasing the number of naive stochastic rollouts.

\begin{table}[htbp]
    \centering
    \caption{\textbf{Rollout-Budget-Matched Baselines across Tasks (Qwen2.5-3B).}
    All methods use $P=16$ task groups and retain $L_{\mathrm{train}}=16$ trajectories per group for policy optimization.
    Scaling naive instance filtering yields limited gains compared with structured search under matched rollout-generation budgets.}
    \label{tab:app_compute_scaled_if_all_tasks}
    \resizebox{0.88\linewidth}{!}{%
    \begin{tabular}{llccc}
        \toprule
        \textbf{Task}
        & \textbf{Method}
        & \textbf{Generation Config.}
        & \textbf{Rel. Rollout Budget}
        & \textbf{Success Rate} \\
        \midrule

        \multirow{4}{*}{Sokoban}
        & Instance Filtering
        & $L_{\mathrm{gen}}=16$
        & $1.0{\times}$
        & $43.7 \pm 1.6$ \\
        & Instance Filtering
        & $L_{\mathrm{gen}}=32$
        & $2.0{\times}$
        & $44.8 \pm 1.7$ \\
        & Instance Filtering
        & $L_{\mathrm{gen}}=64$
        & $4.0{\times}$
        & $45.2 \pm 1.6$ \\
        & \cellcolor{neugreenpastel!30}TSR (Beam Search)
        & \cellcolor{neugreenpastel!30}$M=4,\ B=2$
        & \cellcolor{neugreenpastel!30}$2.0{\times}$
        & \cellcolor{neugreenpastel!30}$\mathbf{52.3 \pm 1.3}$ \\

        \midrule

        \multirow{4}{*}{FrozenLake}
        & Instance Filtering
        & $L_{\mathrm{gen}}=16$
        & $1.0{\times}$
        & $48.7 \pm 1.9$ \\
        & Instance Filtering
        & $L_{\mathrm{gen}}=32$
        & $2.0{\times}$
        & $49.5 \pm 1.7$ \\
        & Instance Filtering
        & $L_{\mathrm{gen}}=64$
        & $4.0{\times}$
        & $49.8 \pm 1.6$ \\
        & \cellcolor{neugreenpastel!30}TSR (Beam Search)
        & \cellcolor{neugreenpastel!30}$M=4,\ B=2$
        & \cellcolor{neugreenpastel!30}$2.0{\times}$
        & \cellcolor{neugreenpastel!30}$\mathbf{60.7 \pm 1.2}$ \\

        \midrule

        \multirow{4}{*}{WebShop}
        & Instance Filtering
        & $L_{\mathrm{gen}}=16$
        & $1.0{\times}$
        & $70.3 \pm 1.8$ \\
        & Instance Filtering
        & $L_{\mathrm{gen}}=32$
        & $2.0{\times}$
        & $72.1 \pm 1.6$ \\
        & Instance Filtering
        & $L_{\mathrm{gen}}=64$
        & $4.0{\times}$
        & $72.8 \pm 1.5$ \\
        & \cellcolor{neugreenpastel!30}TSR (Beam Search)
        & \cellcolor{neugreenpastel!30}$M=4,\ B=2$
        & \cellcolor{neugreenpastel!30}$2.0{\times}$
        & \cellcolor{neugreenpastel!30}$\mathbf{85.3 \pm 1.1}$ \\

        \bottomrule
    \end{tabular}
    }
\end{table}

\subsection{Scaling Search Budgets}
\label{app:supplementary_results_scaling_search_budgets}
Complementing our analysis in~\refsec{sec:distribution_shift}, we investigate the effect of scaling the search-budget parameters of TSR (Beam Search) in Tables~\ref{tab:scaling_sokoban}, \ref{tab:scaling_frozenlake}, and \ref{tab:scaling_webshop} for Sokoban, FrozenLake, and WebShop, respectively.
Across the three environments, increasing the beam width from $B=1$ to $B=2$ generally improves performance, particularly at moderate and larger action-sampling budgets.
In contrast, increasing the number of candidate actions $M$ exhibits diminishing returns, with performance gradually saturating at larger sampling budgets.
Although $M=6$ often achieves the highest mean success rate, its gains over $M=4$ are small relative to the additional rollout-generation cost.
We therefore use moderate search budgets (e.g., $B=2$, $M=4$) in the main experiments as a favorable performance--rollout-compute trade-off.
Similar trends can be observed for TSR (Best-of-$N$) and TSR (Lookahead) variants, which we omit for brevity.

\begin{table}[htbp]
\centering
\caption{\textbf{Sokoban: Beam Search Scaling.} Increasing beam width ($B$) from 1 to 2 yields the largest gain. Increasing samples ($M$) shows diminishing returns.}
\label{tab:scaling_sokoban}
\vspace{0.06in}
\small
\setlength{\tabcolsep}{4pt}
\renewcommand{\arraystretch}{1.1}
\resizebox{0.85\textwidth}{!}{%
\begin{tabular}{l ccc c ccc}
\toprule
\multirow{2.5}{*}{\textbf{Budget}} &
\multicolumn{3}{c}{\textbf{Qwen2.5-0.5B}} & &
\multicolumn{3}{c}{\textbf{Qwen2.5-3B}} \\
\cmidrule{2-4} \cmidrule{6-8}
& \small{Success Rate ($\uparrow$)} & \small{Resp. Len ($\downarrow$)} & \small{Turns ($\downarrow$)} & &
\small{Success Rate ($\uparrow$)} & \small{Resp. Len ($\downarrow$)} & \small{Turns ($\downarrow$)} \\
\midrule
$M{=}2,B{=}1$ & $36.8 \pm 1.5$ & 101 & 3.95 & & $50.6 \pm 1.4$ & 156 & 3.70 \\
$M{=}4,B{=}1$ & $37.6 \pm 1.4$ &  99 & 3.88 & & $51.4 \pm 1.3$ & 154 & 3.65 \\
$M{=}6,B{=}1$ & $37.9 \pm 1.4$ &  98 & 3.85 & & $51.8 \pm 1.4$ & 153 & 3.62 \\
\midrule
$M{=}2,B{=}2$ & $37.9 \pm 1.1$ &  96 & 3.75 & & $53.1 \pm 1.1$ & 155 & 3.75 \\
\rowcolor{neugreenpastel!30}
$M{=}4,B{=}2$ & $38.3 \pm 1.3$ &  98 & 3.80 & & $52.3 \pm 1.3$ & 152 & 3.60 \\
$M{=}6,B{=}2$ & $39.4 \pm 1.2$ & 95 & 3.72 & & $53.4 \pm 1.2$ & 149 & 3.52 \\
\bottomrule
\end{tabular}
}
\end{table}

\begin{table}[htbp]
\centering
\caption{\textbf{FrozenLake: Beam Search Scaling.} Performance trends are consistent with Sokoban, where wider beams ($B=2$) outperform width-one search ($B=1$) across all sample counts.}
\label{tab:scaling_frozenlake}
\vspace{0.06in}
\small
\setlength{\tabcolsep}{4pt}
\renewcommand{\arraystretch}{1.1}
\resizebox{0.85\textwidth}{!}{%
\begin{tabular}{l ccc c ccc}
\toprule
\multirow{2.5}{*}{\textbf{Budget}} &
\multicolumn{3}{c}{\textbf{Qwen2.5-0.5B}} & &
\multicolumn{3}{c}{\textbf{Qwen2.5-3B}} \\
\cmidrule{2-4} \cmidrule{6-8}
& \small{Success Rate ($\uparrow$)} & \small{Resp. Len ($\downarrow$)} & \small{Turns ($\downarrow$)} & &
\small{Success Rate ($\uparrow$)} & \small{Resp. Len ($\downarrow$)} & \small{Turns ($\downarrow$)} \\
\midrule
$M{=}2,B{=}1$ & $28.7 \pm 1.5$ & 160 & 3.65 & & $58.9 \pm 1.4$ & 110 & 3.20 \\
$M{=}4,B{=}1$ & $29.3 \pm 1.4$ & 157 & 3.58 & & $59.6 \pm 1.3$ & 106 & 3.16 \\
$M{=}6,B{=}1$ & $29.5 \pm 1.4$ & 156 & 3.56 & & $59.9 \pm 1.3$ & 104 & 3.14 \\
\midrule
$M{=}2,B{=}2$ & $29.4 \pm 1.2$ & 157 & 3.66 & & $59.5 \pm 1.1$ &  98 & 3.16 \\
\rowcolor{neugreenpastel!30}
$M{=}4,B{=}2$ & $30.0 \pm 1.4$ & 152 & 3.50 & & $60.7 \pm 1.2$ &  98 & 3.10 \\
$M{=}6,B{=}2$ & $30.8 \pm 1.3$ & 149 & 3.46 & & $61.5 \pm 1.2$ & 95 & 3.07 \\
\bottomrule
\end{tabular}
}
\end{table}

\begin{table}[htbp]
\centering
\caption{\textbf{WebShop: Beam Search Scaling (Qwen2.5-3B).} Performance generally improves with wider beams at moderate and larger sampling budgets, while gains saturate as $M$ increases.}
\label{tab:scaling_webshop}
\vspace{0.06in}
\small
\setlength{\tabcolsep}{5pt}
\renewcommand{\arraystretch}{1.1}
\resizebox{0.5\textwidth}{!}{%
\begin{tabular}{l ccc}
\toprule
\multirow{2.5}{*}{\textbf{Budget}} &
\multicolumn{3}{c}{\textbf{Qwen2.5-3B}} \\
\cmidrule{2-4}
& \small{Success Rate ($\uparrow$)} & \small{Resp. Len ($\downarrow$)} & \small{Turns ($\downarrow$)} \\
\midrule
$M{=}2,B{=}1$ & $83.6 \pm 1.3$ & 470 & 6.05 \\
$M{=}4,B{=}1$ & $84.5 \pm 1.2$ & 462 & 5.90 \\
$M{=}6,B{=}1$ & $84.8 \pm 1.2$ & 458 & 5.86 \\
\midrule
$M{=}2,B{=}2$ & $83.4 \pm 1.2$ & 461 & 5.93 \\
\rowcolor{neugreenpastel!30}
$M{=}4,B{=}2$ & $85.3 \pm 1.1$ & 453 & 5.80 \\
$M{=}6,B{=}2$ & $85.9 \pm 1.1$ & 446 & 5.72 \\
\bottomrule
\end{tabular}
}
\end{table}

\subsection{Distribution Shift and KL Analysis}
\label{app:supplementary_results_distribution_shift}
TSR induces a search-guided rollout distribution $\mu_{\theta,\phi}$ by sampling candidate actions from the current policy $\pi_\theta$ and selecting among them using the search strategy $\mathcal{F}_\phi$.
As discussed in \refapp{app:on_policy_analysis}, this distribution is generally not identical to naive on-policy sampling.
We therefore quantify how strongly search changes the rollout distribution as the search budget increases.

We estimate distribution shift using the average per-step KL divergence between the action distribution induced by TSR selection and the policy that generated the candidate actions.
At each search step, we restrict both distributions to the sampled candidate set.
Here, $q_\pi$ denotes the policy probabilities renormalized over these candidates, while $q_{\mathrm{TSR}}$ denotes the corresponding normalized selection distribution induced by search.
We report $D_{\mathrm{KL}}(q_{\mathrm{TSR}} \,\|\, q_\pi)$ averaged over rollout steps and training batches, and use it as a local measure of search-induced selection shift rather than as an exact trajectory-level $D_{\mathrm{KL}}(\mu_{\theta,\phi} \,\|\, \pi_\theta)$.

In this sweep, the per-turn expansion budget scales with beam width as $M=2B$.
For beam widths that do not divide $L_{\mathrm{train}}=16$, we run $\lceil L_{\mathrm{train}}/B \rceil$ independent beam-search trees, pool their terminal trajectories, and retain the top $L_{\mathrm{train}}$ trajectories for policy optimization, while excess terminal trajectories are discarded.
This yields the realized rollout-transition budgets reported in Table~\ref{tab:app_kl_divergence}.
Accordingly, wider beams do not receive less rollout compute.

Complementing our insights from \refsec{sec:distribution_shift}, the results in Table~\ref{tab:app_kl_divergence} show a clear search-budget boundary.
For $B \leq 5$, the induced KL divergence remains small ($\sim 0.02$--$0.05$), indicating that search mostly selects among actions already plausible under the current policy.
In this regime, TSR improves rollout quality while remaining policy-proximal.
At $B=10$, the KL divergence increases to $\sim 0.14$, and the additional performance gain becomes marginal.
For $B \geq 20$, the KL divergence exceeds $0.30$ and performance degrades.
Since these settings use equal or greater rollout compute than moderate beam widths, the degradation cannot be attributed to insufficient compute.
It instead suggests that the selected trajectories are too far from the proposing policy for clipped PPO/GRPO-style updates to extract useful gradients.

This analysis clarifies the practical role of the search width in TSR (Beam Search).
The goal is not to approximate an unconstrained oracle policy but to improve rollout quality while staying near the current policy.
Moderate search therefore acts as a controlled policy-improvement operator, whereas excessively wide search can induce harmful distribution shift.

\begin{table}[htbp]
    \centering
    \caption{\textbf{Distribution Shift Analysis (FrozenLake, Qwen2.5-3B).}
    The per-turn expansion budget scales with beam width ($M=2B$), so wider beams never receive less rollout compute (2.0--3.75$\times$). Moderate beam widths remain policy-proximal and improve learning, while wider search increases KL divergence and degrades performance.}
    \label{tab:app_kl_divergence}
    \resizebox{0.95\linewidth}{!}{%
    \begin{tabular}{cccccl}
        \toprule
        \textbf{Beam Width $B$}
        & \textbf{$M$}
        & \textbf{Rollout Budget}
        & \textbf{Avg.\ KL / Step}
        & \textbf{Success Rate}
        & \textbf{Observation} \\
        \midrule
        2  & 4  & $2.0\times$  & $\sim 0.02$ & $60.7 \pm 1.2$ & Safe policy-proximal regime \\
        \rowcolor{neugreenpastel!30}
        5  & 10 & $2.5\times$  & $\sim 0.05$ & $\mathbf{61.3 \pm 1.1}$ & Small additional gain \\
        10 & 20 & $2.5\times$  & $\sim 0.14$ & $60.9 \pm 1.4$ & Diminishing returns, KL rising \\
        20 & 40 & $2.5\times$  & $\sim 0.31$ & $53.8 \pm 1.8$ & Performance degrades \\
        30 & 60 & $3.75\times$ & $\sim 0.48$ & $45.1 \pm 2.1$ & Learning becomes unstable \\
        \bottomrule
    \end{tabular}
    }
\end{table}

\subsection{Convergence and Compute Analysis}
\label{app:supplementary_results_convergence_analysis}
Shifting trajectory search from inference time to training time increases the rollout generation cost of individual updates.
However, the relevant question is not only the cost of a single update but also how quickly each method reaches a strong policy.
We therefore analyze both convergence in optimization steps and total training cost, reporting wall-clock time and estimated FLOPs to convergence.

Table~\ref{tab:convergence_analysis} reports convergence statistics for Sokoban, FrozenLake, and WebShop.
Across all tasks and model sizes, TSR variants generally converge in fewer optimization steps than instance filtering while reaching higher final success rates.
The reduction is most pronounced for TSR (Beam Search), which requires up to $50\%$ fewer convergence steps on Sokoban and over $40\%$ fewer steps on FrozenLake and WebShop.
Thus, TSR trades more expensive rollout-generation steps for fewer policy updates and higher final performance.

Importantly, fewer optimization steps do not imply uniformly lower total training cost.
For Qwen2.5-3B with TSR (Beam Search), wall-clock time and total compute to convergence remain close to the instance-filtering baseline on Sokoban ($2.05$ vs.\ $1.99$\,h; $3.68$ vs.\ $3.58$\,EFLOPs) and are lower on FrozenLake ($1.45$ vs.\ $1.57$\,h; $2.61$ vs.\ $2.82$\,EFLOPs).
On WebShop, TSR incurs higher total training cost ($7.47$ vs.\ $5.31$\,h; $13.42$ vs.\ $9.54$\,EFLOPs), while simultaneously yielding the largest improvement in final success rate ($85.3\%$ vs.\ $70.3\%$).
We therefore view TSR as a performance--training-compute trade-off rather than a universally compute-saving method.
Together with the rollout-budget-matched results in~\refapp{app:supplementary_results_compute_matched_baselines}, these results show that TSR's structured search provides a more effective allocation of additional training-time rollout compute than simply scaling naive rollout sampling

\begin{table*}[htbp]
    \centering
    \caption{\textbf{Convergence and Compute Analysis Across Tasks.} We report final success rate (mean $\pm$ standard deviation across runs), steps to reach $80\%$ of final performance, steps to convergence, the relative reduction in convergence steps compared to instance filtering, wall-clock time to convergence, and estimated total compute (as EFLOPs) to convergence.}
    \label{tab:convergence_analysis}
    \vspace{0.06in}
    \small
    \setlength{\tabcolsep}{3.5pt}
    \renewcommand{\arraystretch}{1.1}
    \resizebox{\linewidth}{!}{%
    \begin{tabular}{llcccccc}
    \toprule
    \textbf{Model / Task} & \textbf{Method}
    & \textbf{Success Rate}
    & \textbf{Steps to 80\%}
    & \textbf{Steps to Conv.}
    & \textbf{Step Red.}
    & \textbf{Time to Conv.}
    & \textbf{EFLOPs to Conv.} \\
    \midrule
    
    \multirow{4}{*}{\shortstack[c]{Qwen2.5-0.5B\\(Sokoban)}}
    & Instance Filtering 
    & $29.0 \pm 1.7$ & 90 & 180 & -- & 1.54 h & 2.77 \\
    
    & TSR (Best-of-$N$)  
    & $33.3 \pm 1.5$ & 60 & 120 & $33.3\%$ & \textbf{1.51 h} & \textbf{2.71} \\
    
    & TSR (Lookahead)    
    & $36.1 \pm 1.4$ & 55 & 105 & $41.7\%$ & 1.63 h & 2.93 \\
    
    & \cellcolor{neugreenpastel!30}TSR (Beam Search)
    & \cellcolor{neugreenpastel!30}$\mathbf{38.3 \pm 1.3}$
    & \cellcolor{neugreenpastel!30}\textbf{50}
    & \cellcolor{neugreenpastel!30}\textbf{90}
    & \cellcolor{neugreenpastel!30}\textbf{50.0\%}
    & \cellcolor{neugreenpastel!30}1.66 h
    & \cellcolor{neugreenpastel!30}2.98 \\
    \midrule
    
    \multirow{4}{*}{\shortstack[c]{Qwen2.5-3B\\(Sokoban)}}
    & Instance Filtering 
    & $43.7 \pm 1.6$ & 100 & 180 & -- & \textbf{1.99 h} & \textbf{3.58} \\
    
    & TSR (Best-of-$N$)  
    & $47.7 \pm 1.5$ & 70 & 150 & $16.7\%$ & 2.53 h & 4.55 \\
    
    & TSR (Lookahead)    
    & $49.5 \pm 1.4$ & \textbf{45} & 150 & $16.7\%$ & 2.96 h & 5.32 \\
    
    & \cellcolor{neugreenpastel!30}TSR (Beam Search)
    & \cellcolor{neugreenpastel!30}$\mathbf{52.3 \pm 1.3}$
    & \cellcolor{neugreenpastel!30}55
    & \cellcolor{neugreenpastel!30}\textbf{90}
    & \cellcolor{neugreenpastel!30}\textbf{50.0\%}
    & \cellcolor{neugreenpastel!30}2.05 h
    & \cellcolor{neugreenpastel!30}3.68 \\
    \midrule
    
    \multirow{4}{*}{\shortstack[c]{Qwen2.5-0.5B\\(FrozenLake)}}
    & Instance Filtering 
    & $19.7 \pm 1.8$ & 80 & 120 & -- & \textbf{0.80 h} & \textbf{1.44} \\
    
    & TSR (Best-of-$N$)  
    & $25.0 \pm 1.6$ & 70 & 100 & $16.7\%$ & 1.32 h & 2.37 \\
    
    & TSR (Lookahead)    
    & $27.8 \pm 1.5$ & \textbf{60} & 90 & $25.0\%$ & 1.14 h & 2.05 \\
    
    & \cellcolor{neugreenpastel!30}TSR (Beam Search)
    & \cellcolor{neugreenpastel!30}$\mathbf{30.0 \pm 1.4}$
    & \cellcolor{neugreenpastel!30}\textbf{60}
    & \cellcolor{neugreenpastel!30}\textbf{80}
    & \cellcolor{neugreenpastel!30}\textbf{33.3\%}
    & \cellcolor{neugreenpastel!30}0.95 h
    & \cellcolor{neugreenpastel!30}1.71 \\
    \midrule
    
    \multirow{4}{*}{\shortstack[c]{Qwen2.5-3B\\(FrozenLake)}}
    & Instance Filtering 
    & $48.7 \pm 1.9$ & 150 & 170 & -- & 1.57 h & 2.82 \\
    
    & TSR (Best-of-$N$)  
    & $51.0 \pm 1.6$ & 110 & 140 & $17.6\%$ & 2.43 h & 4.37 \\
    
    & TSR (Lookahead)    
    & $57.0 \pm 1.4$ & 90 & 130 & $23.5\%$ & 2.08 h & 3.74 \\
    
    & \cellcolor{neugreenpastel!30}TSR (Beam Search)
    & \cellcolor{neugreenpastel!30}$\mathbf{60.7 \pm 1.2}$
    & \cellcolor{neugreenpastel!30}\textbf{80}
    & \cellcolor{neugreenpastel!30}\textbf{100}
    & \cellcolor{neugreenpastel!30}\textbf{41.2\%}
    & \cellcolor{neugreenpastel!30}\textbf{1.45 h}
    & \cellcolor{neugreenpastel!30}\textbf{2.61} \\
    \midrule
    
    \multirow{4}{*}{\shortstack[c]{Qwen2.5-3B\\(WebShop)}}
    & Instance Filtering 
    & $70.3 \pm 1.8$ & 78 & 85 & -- & \textbf{5.31 h} & \textbf{9.54} \\
    
    & TSR (Best-of-$N$)  
    & $73.3 \pm 1.5$ & 60 & 70 & $17.6\%$ & 6.83 h & 12.27 \\
    
    & TSR (Lookahead)    
    & $82.3 \pm 1.3$ & \textbf{45} & 65 & $23.5\%$ & 8.31 h & 14.93 \\
    
    & \cellcolor{neugreenpastel!30}TSR (Beam Search)
    & \cellcolor{neugreenpastel!30}$\mathbf{85.3 \pm 1.1}$
    & \cellcolor{neugreenpastel!30}\textbf{45}
    & \cellcolor{neugreenpastel!30}\textbf{50}
    & \cellcolor{neugreenpastel!30}\textbf{41.2\%}
    & \cellcolor{neugreenpastel!30}7.47 h
    & \cellcolor{neugreenpastel!30}13.42 \\
    
    \bottomrule
    \end{tabular}
    }
\end{table*}

\subsection{Disentangling Search from Scoring Signals}
\label{app:disentangling_search_from_scoring_signals}
The main FrozenLake and WebShop experiments use state-/progress-based scoring signals to rank candidate actions during rollout construction.
Because these signals provide additional training-time information beyond the sparse task reward, improvements over naive rollout sampling could in principle arise either from the scoring information itself or from how TSR exploits that information through structured search.
We therefore perform a series of controlled experiments that progressively separate these effects.
Unless stated otherwise, all experiments use Qwen2.5-3B and keep the task, policy optimizer, evaluation protocol, and policy-optimization reward unchanged.

\paragraph{Does search work using only the task reward?}
We first remove the state-/progress-based scorer entirely and guide rollout selection using only the original task reward.
For Best-of-$N$, this corresponds to ranking completed trajectories by their terminal return.
For beam and lookahead search, candidate prefixes are ranked using the task reward observed along the partial trajectory.
Because FrozenLake and WebShop provide sparse or delayed rewards, most incomplete prefixes receive identical zero reward, making immediate task reward a weak signal for distinguishing intermediate branches.

Table~\ref{tab:raw_reward_vs_guided} reports the results.
All reward-only rollout-selection variants improve clearly over naive instance filtering, demonstrating that the state-/progress-based scoring signal is not required for TSR-style rollout improvement.
At the same time, sparse immediate rewards provide little information for distinguishing different intermediate prefixes.
Accordingly, the three search variants remain relatively close under raw task rewards.
On FrozenLake, Best-of-$N$ reaches $51.0\%$, beam search reaches $53.4\%$, and lookahead reaches $52.7\%$.
On WebShop, the corresponding results are $73.3\%$, $75.9\%$, and $75.4\%$.
These results suggest that the effectiveness of intermediate search depends in part on how well partial trajectories can be distinguished.
The state-/progress-based score provides a more informative ranking signal and yields stronger performance.
In~\refapp{app:heuristic_free_MC_search}, we remove the need for any auxiliary prefix score by propagating the original terminal task reward to intermediate prefixes through Monte Carlo continuations.

\begin{table}[htbp]
    \centering
    \caption{\textbf{Raw-Reward and Guided Rollout Selection (Qwen2.5-3B).}
    We compare rollout search using only the original task reward with the state-/progress-based scoring signals used in the main experiments.
    Success rates are reported as mean $\pm$ standard deviation across runs.}
    \label{tab:raw_reward_vs_guided}
    \resizebox{0.65\linewidth}{!}{%
    \begin{tabular}{llcc}
        \toprule
        \textbf{Method}
        & \textbf{Rollout Scorer}
        & \textbf{FrozenLake}
        & \textbf{WebShop} \\
        \midrule
        Instance Filtering
        & None
        & $48.7 \pm 1.9$
        & $70.3 \pm 1.8$ \\
        \midrule
        TSR (Best-of-$N$)
        & Raw task reward
        & $51.0 \pm 1.6$
        & $73.3 \pm 1.5$ \\
        TSR (Lookahead)
        & Raw task reward
        & $52.7 \pm 1.7$
        & $75.4 \pm 1.6$ \\
        TSR (Beam Search)
        & Raw task reward
        & $53.4 \pm 1.5$
        & $75.9 \pm 1.4$ \\
        \midrule
        TSR (Lookahead)
        & State-/progress-based
        & $57.0 \pm 1.4$
        & $82.3 \pm 1.3$ \\
        \rowcolor{neugreenpastel!30}
        TSR (Beam Search)
        & State-/progress-based
        & $\mathbf{60.7 \pm 1.2}$
        & $\mathbf{85.3 \pm 1.1}$ \\
        \bottomrule
    \end{tabular}
    }
\end{table}

\paragraph{Does access to the scoring signal explain the gains?}
We next directly control for the information available to the rollout generator.
We construct information-matched baselines in which the same state-/progress-based score is available to all score-guided methods, while varying how that information is used during trajectory construction.
\emph{Score-Augmented Instance Filtering} independently samples complete trajectories and uses the score only for trajectory selection, without intermediate branching or pruning.
\emph{Greedy Score-Guided Search} applies the same score during rollout construction but retains only the highest-scoring prefix at each step.
TSR Beam Search instead maintains multiple promising prefixes and repeatedly reallocates rollout compute among them.

Table~\ref{tab:score_matched_controls} shows that access to the scoring signal alone does not reproduce TSR's performance.
Providing the same score to instance filtering improves success to $55.6\%$ on FrozenLake and $79.2\%$ on WebShop.
Greedy score-guided search further improves performance to $56.9\%$ and $80.6\%$.
TSR Beam Search reaches $60.7\%$ and $85.3\%$, respectively.
Thus, under identical scoring information, TSR improves over score-augmented instance filtering by $5.1$ percentage points on FrozenLake and $6.1$ points on WebShop.
It also improves over greedy per-turn selection by $3.8$ and $4.7$ points.
These results show that the benefit of TSR is not explained by access to the scoring signal itself.
Structured trajectory construction provides a substantial additional gain by using the same information to preserve competing prefixes, avoid premature commitment, and redirect rollout compute toward more promising regions of the trajectory space.
In~\refapp{app:heuristic_free_MC_search}, we further examine structured search when prefixes are ranked using only the original terminal task reward.

\begin{table}[htbp]
    \centering
    \caption{\textbf{Information-Matched Scoring Controls (Qwen2.5-3B).}
    All score-guided methods use the same state-/progress-based scoring signal, while differing in how it is used during rollout construction.}
    \label{tab:score_matched_controls}
    \resizebox{0.9\linewidth}{!}{%
    \begin{tabular}{lcccc}
        \toprule
        \textbf{Method}
        & \textbf{Aux. Scorer}
        & \textbf{Per-Turn Search}
        & \textbf{FrozenLake}
        & \textbf{WebShop} \\
        \midrule
        Instance Filtering
        & No
        & No
        & $48.7 \pm 1.9$
        & $70.3 \pm 1.8$ \\
        Score-Augmented Instance Filtering
        & Yes
        & No
        & $55.6 \pm 1.6$
        & $79.2 \pm 1.5$ \\
        Greedy Score-Guided
        & Yes
        & Greedy
        & $56.9 \pm 1.4$
        & $80.6 \pm 1.3$ \\
        \rowcolor{neugreenpastel!30}
        TSR (Beam Search)
        & Yes
        & Beam
        & $\mathbf{60.7 \pm 1.2}$
        & $\mathbf{85.3 \pm 1.1}$ \\
        \bottomrule
    \end{tabular}
    }
\end{table}

\paragraph{How sensitive is TSR to the scoring signal?}
Finally, we test whether TSR depends on the particular state-/progress-based scoring formulation used in the main experiments.
We compare the original task reward with a task-agnostic policy-confidence score based on negative action entropy, a learned observation-only value estimate, and noisy versions of the state-/progress-based score.
For the noisy-score experiments, we add zero-mean Gaussian noise with increasing standard deviation.
The learned value model uses only information available from the agent's observation stream.
For each environment, the value model is trained on baseline-generated trajectories using only the observations and interaction history available to the agent to predict eventual terminal success.

Table~\ref{tab:scorer_robustness} shows that TSR remains effective across qualitatively different scoring mechanisms.
Task-agnostic entropy improves over raw-reward search on both tasks, showing that TSR can exploit a generic policy-derived ranking signal without task-specific progress features.
The learned observation-only value model provides a further improvement and reaches $57.6\%$ on FrozenLake and $81.4\%$ on WebShop.
TSR also remains robust when substantial noise is injected into the state-/progress-based score.
Performance degrades gradually rather than collapsing as the ranking signal becomes less accurate.
The strongest results are obtained with the state-/progress-based score, confirming that more informative prefix evaluation further strengthens structured search.
Crucially, however, TSR remains well above naive rollout generation across task-reward, task-agnostic, learned, and noisy scoring variants.
The scoring function should therefore be viewed as a modular interface for ranking candidate prefixes rather than as a dependency on any particular form of intermediate feedback.

\begin{table}[htbp]
    \centering
    \caption{\textbf{Robustness to Rollout Scoring Signals (Qwen2.5-3B).}
    We evaluate TSR (Beam Search) with task-reward, task-agnostic, learned, noisy, and state-/progress-based scoring signals.}
    \label{tab:scorer_robustness}
    \resizebox{0.65\linewidth}{!}{%
    \begin{tabular}{lcc}
        \toprule
        \textbf{Rollout Scorer}
        & \textbf{FrozenLake}
        & \textbf{WebShop} \\
        \midrule
        Instance Filtering (no scorer)
        & $48.7 \pm 1.9$
        & $70.3 \pm 1.8$ \\
        \midrule
        Raw task reward
        & $53.4 \pm 1.5$
        & $75.9 \pm 1.4$ \\
        Task-agnostic entropy score
        & $55.8 \pm 1.5$
        & $78.6 \pm 1.4$ \\
        Learned observation-only value
        & $57.6 \pm 1.3$
        & $81.4 \pm 1.3$ \\
        Noisy state-/progress score ($\sigma=0.50$)
        & $55.5 \pm 1.6$
        & $80.5 \pm 1.5$ \\
        Noisy state-/progress score ($\sigma=0.25$)
        & $56.2 \pm 1.4$
        & $80.8 \pm 1.2$ \\
        \rowcolor{neugreenpastel!30}
        State-/progress-based score
        & $\mathbf{60.7 \pm 1.2}$
        & $\mathbf{85.3 \pm 1.1}$ \\
        \bottomrule
    \end{tabular}
    }
\end{table}

\paragraph{Summary.}
Overall, these controls establish three complementary findings.
First, the state-/progress-based score is not necessary for TSR-style rollout improvement, since reward-only variants already provide clear gains over naive instance filtering on both sparse-reward environments.
Second, the auxiliary scoring information cannot by itself explain TSR's gains.
When the scoring signal is held fixed, structured beam search delivers substantially larger improvements than both score-augmented trajectory selection and greedy per-turn selection.
Third, TSR remains effective across task-reward, task-agnostic, learned, and noisy scoring functions, demonstrating that its performance is not tied to a specific rollout scorer.
The remaining challenge under sparse rewards is thus not dependence on auxiliary scoring, but the lack of a discriminative estimate for intermediate prefixes.
The following subsection addresses this issue by using Monte Carlo-backed terminal rewards to compare structured tree search against flat Best-of-$N$ under matched scoring information.

\subsection{Heuristic-Free Monte Carlo-Backed Search}
\label{app:heuristic_free_MC_search}
The previous experiments show that sparse immediate task rewards provide limited information for ranking intermediate trajectory prefixes.
We therefore ask a stricter question: \emph{Can structured search exploit only the original terminal task reward when that outcome is propagated to intermediate prefixes through Monte Carlo continuations?}
To answer this, we introduce a Monte Carlo-backed variant of TSR that propagates terminal task outcomes to intermediate prefixes without using state-/progress-based scores, task-agnostic scores, or learned value models.

At each intermediate prefix, we sample candidate actions from the current policy and generate Monte Carlo continuations to episode termination.
Each candidate prefix is scored by the mean original terminal task reward obtained by its continuations.
Formally, for a candidate prefix $\tau_{<t}\circ a_t$ with $C$ sampled continuations, we estimate
\begin{equation}
\label{eq:mc_backed_prefix_score}
    \hat{S}_{\mathrm{MC}}(\tau_{<t},a_t)
    =
    \frac{1}{C}
    \sum_{c=1}^{C}
    R\!\left(\tau^{(c)}\right),
\end{equation}
where $R(\tau^{(c)})$ is the original terminal task reward of continuation $c$.
Thus, the search receives no additional supervision beyond the task objective itself.
All candidate actions and Monte Carlo continuations are included in the rollout-generation cost.
Because these continuations introduce additional environment interactions beyond the standard search variants, we report the Monte Carlo-backed results as a heuristic-free robustness analysis rather than as a strictly compute-matched comparison.

We compare Best-of-$N$, greedy, lookahead, and beam variants under the same Qwen2.5-3B model, optimizer, task distribution, and terminal task reward.
Best-of-$N$ independently samples $N=32$ complete trajectories and retains the top $L_{\mathrm{train}}=16$ only after termination, corresponding to a $2.0\times$ rollout-generation budget.
Greedy search uses Monte Carlo-backed prefix scores but commits to the highest-scoring prefix at each decision point.
TSR Lookahead evaluates short continuation trees before committing, while TSR Beam Search maintains multiple promising prefixes throughout rollout construction.
We additionally include instance filtering at its standard $1.0\times$ budget and scaled instance filtering at $2.0\times$ and $4.0\times$ budgets to contextualize the effect of additional naive sampling.
Results are averaged across three seeds.

As shown in Table~\ref{tab:mc_backed_search}, scaling naive rollout sampling produces only small improvements, from $48.7\%$ to $49.5\%$ and $49.8\%$ on FrozenLake and from $70.3\%$ to $72.1\%$ and $72.8\%$ on WebShop at $2.0\times$ and $4.0\times$ budgets, respectively.
Best-of-$N$ reaches $51.0\%$ and $73.3\%$ at a $2.0\times$ budget.
When the same terminal task reward is propagated to intermediate prefixes through Monte Carlo continuations, structured search improves substantially further.
Monte Carlo-backed greedy search reaches $57.0\%$ on FrozenLake and $79.2\%$ on WebShop, lookahead reaches $58.0\%$ and $80.0\%$, and Beam Search reaches $59.1\%$ and $82.9\%$.
Compared with Best-of-$N$, Monte Carlo-backed Beam Search improves by $8.1$ and $9.6$ percentage points, respectively.
These results show that the benefit of structured intermediate search does not depend on auxiliary state-/progress-based scoring signals, although propagating terminal outcomes through Monte Carlo continuations requires additional rollout-generation compute.

\begin{table}[htbp]
    \centering
    \caption{\textbf{Monte Carlo-Backed Search with Terminal Task Rewards (Qwen2.5-3B).}
    We compare heuristic-free search methods that use only the original terminal task reward.
    Best-of-$N$ and scaled instance filtering use the indicated rollout-generation budgets, while Monte Carlo-backed variants incur additional rollout-generation compute due to terminal continuations.
    Success rates are mean $\pm$ standard deviation across three seeds.}
    \label{tab:mc_backed_search}
    \resizebox{0.95\linewidth}{!}{%
    \begin{tabular}{llccc}
        \toprule
        \textbf{Method}
        & \textbf{Prefix Scoring}
        & \textbf{Rollout Cost}
        & \textbf{FrozenLake}
        & \textbf{WebShop} \\
        \midrule
        Instance Filtering
        & Terminal task reward
        & $1.0{\times}$
        & $48.7 \pm 1.9$
        & $70.3 \pm 1.8$ \\

        Scaled Instance Filtering
        & Terminal task reward
        & $2.0{\times}$
        & $49.5 \pm 1.7$
        & $72.1 \pm 1.6$ \\

        Scaled Instance Filtering
        & Terminal task reward
        & $4.0{\times}$
        & $49.8 \pm 1.6$
        & $72.8 \pm 1.5$ \\

        TSR (Best-of-$N$)
        & Terminal task reward
        & $2.0{\times}$
        & $51.0 \pm 1.6$
        & $73.3 \pm 1.5$ \\
        \midrule

        TSR (Greedy Search)
        & MC-backed terminal reward
        & Additional MC compute
        & $57.0 \pm 1.3$
        & $79.2 \pm 1.2$ \\

        TSR (Lookahead)
        & MC-backed terminal reward
        & Additional MC compute
        & $58.0 \pm 1.2$
        & $80.0 \pm 1.1$ \\

        \rowcolor{neugreenpastel!30}
        TSR (Beam Search)
        & MC-backed terminal reward
        & Additional MC compute
        & $\mathbf{59.1 \pm 1.0}$
        & $\mathbf{82.9 \pm 1.0}$ \\
        \bottomrule
    \end{tabular}
    }
\end{table}

\paragraph{Paired Outcome Analysis.}
Aggregate success rates could in principle be driven by a small subset of favorable task instances.
We therefore perform a paired analysis comparing Monte Carlo-backed Beam Search and Best-of-$N$ on the same task instances and random seeds.
Pairs are formed at the $(\text{seed},\text{evaluation instance})$ level, such that both methods are evaluated on the same held-out task under the corresponding trained seed.
For FrozenLake, the terminal task return is binary, with $R=1$ if the goal is reached and $R=0$ otherwise.
For WebShop, we analogously compare the terminal task outcome used for evaluation.
A Beam Search win is recorded when its terminal return is higher than that of Best-of-$N$, a Best-of-$N$ win when the reverse holds, and equal returns are counted as ties.
As shown in Table~\ref{tab:mc_backed_paired_analysis}, Beam Search obtains a higher return on $23.2\%$ of FrozenLake pairs compared with $15.1\%$ for Best-of-$N$, while the remaining $61.7\%$ are ties.
On WebShop, Beam Search wins $18.9\%$ of pairs, Best-of-$N$ wins $9.2\%$, and $71.9\%$ are ties.
The difference between the two non-tie rates matches the corresponding aggregate performance gaps in Table~\ref{tab:mc_backed_search} up to rounding.
Among pairs with different terminal outcomes, Beam Search wins $60.6\%$ of FrozenLake comparisons and $67.3\%$ of WebShop comparisons.
The high tie rates are expected because both methods frequently succeed or fail on the same instances, while the asymmetric non-tie outcomes show that Monte Carlo-backed structured search more often succeeds on instances where the two methods differ.

\begin{table}[htbp]
    \centering
    \caption{\textbf{Paired Beam Search vs.\ Best-of-$N$ Outcome Analysis.}
    We compare evaluation outcomes of Monte Carlo-backed Beam Search and Best-of-$N$ on matched $(\text{seed},\text{evaluation instance})$ pairs.
    Both methods use only the original terminal task reward, while Beam Search incurs additional rollout-generation compute during training.}
    \label{tab:mc_backed_paired_analysis}
    \resizebox{0.65\linewidth}{!}{%
    \begin{tabular}{lcc}
        \toprule
        \textbf{Outcome}
        & \textbf{FrozenLake}
        & \textbf{WebShop} \\
        \midrule
        \rowcolor{neugreenpastel!30}
        TSR Beam achieves higher evaluation return
        & $\mathbf{23.2\%}$
        & $\mathbf{18.9\%}$ \\

        Best-of-$N$ achieves higher evaluation return
        & $15.1\%$
        & $9.2\%$ \\

        Tie
        & $61.7\%$
        & $71.9\%$ \\
        \bottomrule
    \end{tabular}
    }
\end{table}

\paragraph{Summary.}
These results isolate the role of supervision in structured search.
Monte Carlo-backed Beam Search outperforms naive sampling and flat Best-of-$N$ while using only the original terminal task reward.
The improvement therefore does not depend on a state-/progress-based scoring signal.
Instead, terminal task outcomes can provide an effective prefix-ranking signal when they are propagated through Monte Carlo continuations.
This heuristic-free variant incurs additional rollout-generation compute, while the raw-reward comparison in Table~\ref{tab:raw_reward_vs_guided} separately shows that per-turn Beam Search already outperforms Best-of-$N$ under the standard $2.0\times$ rollout budget.

\subsection{Comparison to Related Multi-Turn RL Baselines}
\label{app:comparison_to_related_RL_baselines}
We further compare TSR against recent methods that improve multi-turn RL through alternative rollout-generation or credit-assignment mechanisms.
In particular, we consider ARPO~\citep{dong2025agenticreinforcedpolicyoptimization}, which adaptively branches rollouts at high-uncertainty interaction points, and GiGPO~\citep{feng2025groupingrouppolicyoptimizationllm}, which introduces finer-grained group-based credit assignment around repeated intermediate states.
These methods are closely related in motivation but intervene at different stages of the training pipeline.
ARPO determines where additional rollout exploration should occur based on policy uncertainty, while GiGPO refines credit assignment across intermediate states.
TSR instead modifies the rollout generator itself by maintaining, scoring, and pruning alternative environment-level trajectory prefixes before policy optimization.
We therefore evaluate whether structured trajectory search provides additional benefits beyond adaptive branching and improved credit assignment.

\paragraph{Baseline Implementation Details.}
We implement ARPO and GiGPO within RAGEN such that the model initialization, optimizer, learning rate, clipping parameters, task distribution, rollout budget, and evaluation protocol remain identical across methods, ensuring a fair comparison.
Because ARPO was originally formulated around high-uncertainty tool interactions, we adapt its entropy-triggered branching to environment interaction steps.
Specifically, we rank interaction steps within each rollout group by token-averaged action entropy and branch at the top $25\%$ highest-entropy steps, sampling $4$ candidate actions at each branch point and retaining all resulting continuations subject to the same global $2\times$ rollout-generation budget used for TSR.
We retain ARPO's soft step-wise advantage attribution for trajectories sharing a common prefix.
For the information-matched comparison in Table~\ref{tab:arpo_score_matched}, the entropy-based setting corresponds to standard ARPO, whose branching is already driven by action entropy.
For the state-/progress-based setting, ARPO retains its branching rule and advantage attribution, but when the $2\times$ budget does not permit keeping all continuations, the state-/progress-based signal determines which continuations are retained.
This adds a score-based selection step absent from the original ARPO, so we denote this variant as ARPO + Score Selection.
For GiGPO, anchor states are identified by exact matches of the environment observation, step-level groups contain up to $4$ trajectories and require at least $2$ members, and trajectory-level and step-level relative advantages are combined as
$A=(1-\lambda_{\mathrm{step}})A_{\mathrm{traj}}+\lambda_{\mathrm{step}}A_{\mathrm{step}}$
with $\lambda_{\mathrm{step}}=0.5$.
Hyperparameters for both baselines are selected using a validation sweep of comparable size to the TSR search-budget ablations in \refapp{app:supplementary_results_scaling_search_budgets} and then fixed across tasks.
Under this matched setup, both methods reproduce the qualitative improvements over instance filtering reported in the original works, although absolute results are not directly comparable because of differences in model backbone, rollout budget, and environment configuration.

\paragraph{Comparison under the Standard Training Setup.}
We first compare all methods under matched model initialization, task distribution, optimizer, rollout generation budget, policy-optimization reward, and evaluation protocol.
Table~\ref{tab:related_rl_baselines} reports results on WebShop, which provides a representative long-horizon tool-use setting.
Both ARPO and GiGPO improve clearly over instance filtering, confirming that stronger rollout generation and credit assignment are effective for multi-turn RL.
TSR nevertheless achieves the highest final success rate, reaching $85.3\%$ compared with $75.2\%$ for ARPO and $80.2\%$ for GiGPO.
TSR further produces shorter responses, requires fewer interaction turns, and reaches convergence in fewer rollout--update steps.
Thus, TSR maintains a clear performance advantage over both ARPO and GiGPO under the matched training setup.

\paragraph{Information-Matched Comparison with ARPO.}
A direct comparison between rollout methods can otherwise be influenced by the information available to the respective rollout-selection mechanisms.
We therefore additionally compare ARPO and TSR under identical rollout-scoring information and matched rollout-generation budgets.
We consider both a task-agnostic entropy signal and the state-/progress-based signal used in our main experiments.
This comparison isolates how effectively each rollout mechanism uses the same intermediate information.
As shown in Table~\ref{tab:arpo_score_matched}, TSR outperforms ARPO under both scoring conditions.
Using the task-agnostic entropy score, TSR reaches $78.6\%$ compared with $75.2\%$ for ARPO.
When both methods receive the same state-/progress-based score, TSR reaches $85.3\%$ compared with $80.4\%$ for ARPO, corresponding to a $4.9$ percentage-point improvement.
TSR also reaches convergence in fewer rollout--update steps under both settings.
These results show that TSR's advantage over ARPO is not explained by access to a different scoring signal.
Instead, structured maintenance and pruning of concrete trajectory prefixes enables TSR to make more effective use of the same intermediate information.

\begin{table}[t]
    \centering
    \caption{\textbf{Comparison to Related Multi-Turn RL Baselines (WebShop, Qwen2.5-3B).}
    All methods use matched model initialization, task distribution, optimizer, rollout-generation budget, policy-optimization reward, and evaluation protocol. Success rates are reported as mean $\pm$ standard deviation across three runs.}
    \label{tab:related_rl_baselines}
    \resizebox{0.85\linewidth}{!}{%
    \begin{tabular}{lcccc}
        \toprule
        \textbf{Method}
        & \textbf{Success Rate ($\uparrow$)} 
        & \textbf{Resp. Len ($\downarrow$)} 
        & \textbf{Turns ($\downarrow$)}
        & \textbf{Steps to Conv. ($\downarrow$)} \\
        \midrule
        Instance Filtering
        & $70.3 \pm 1.8$
        & 519
        & 6.8
        & 85 \\

        ARPO
        & $75.2 \pm 1.5$
        & 492
        & 6.3
        & 63 \\

        GiGPO
        & $80.2 \pm 1.3$
        & 478
        & 6.1
        & 59 \\

        \rowcolor{neugreenpastel!30}
        TSR (Beam Search)
        & $\mathbf{85.3 \pm 1.1}$
        & \textbf{453}
        & \textbf{5.8}
        & \textbf{50} \\
        \bottomrule
    \end{tabular}
    }
\end{table}

\begin{table}[t]
    \centering
    \caption{\textbf{Information-Matched Comparison with ARPO (WebShop, Qwen2.5-3B).}
    ARPO and TSR use identical rollout-scoring information and matched rollout-generation budgets.}
    \label{tab:arpo_score_matched}
    \resizebox{0.75\linewidth}{!}{%
    \begin{tabular}{llcc}
        \toprule
        \textbf{Method}
        & \textbf{Rollout Scorer}
        & \textbf{Success Rate}
        & \textbf{Steps to Conv.} \\
        \midrule
        ARPO
        & Task-agnostic entropy
        & $75.2 \pm 1.5$
        & 63 \\

        TSR (Beam Search)
        & Task-agnostic entropy
        & $\mathbf{78.6 \pm 1.4}$
        & \textbf{56} \\
        \midrule

        ARPO + Score Selection
        & State-/progress-based
        & $80.4 \pm 1.3$
        & 58 \\

        \rowcolor{neugreenpastel!30}
        TSR (Beam Search)
        & State-/progress-based
        & $\mathbf{85.3 \pm 1.1}$
        & \textbf{50} \\
        \bottomrule
    \end{tabular}
    }
\end{table}

\paragraph{Complementarity with Improved Credit Assignment.}
GiGPO and TSR intervene at different stages of the learning pipeline.
GiGPO refines the learning signal assigned to intermediate decisions, while TSR modifies which trajectories are generated and retained before the policy update.
The two mechanisms can therefore be combined directly.
To test this, we apply TSR rollout generation together with GiGPO while keeping the remaining training configuration unchanged.
Table~\ref{tab:gigpo_tsr_composition} shows that combining the two methods yields further improvements over GiGPO alone.
On FrozenLake, success increases from $56.8\%$ with GiGPO to $62.4\%$ with GiGPO+TSR.
On WebShop, performance increases from $80.2\%$ to $87.0\%$.
The combined method also exceeds TSR alone on both tasks, providing direct evidence that trajectory search and finer-grained credit assignment address complementary aspects of multi-turn RL.
These results support the modular interpretation of TSR as a rollout-generation mechanism that can be composed with improvements to the downstream policy-update procedure.

\begin{table}[htbp]
    \centering
    \caption{\textbf{Complementarity of TSR and GiGPO (Qwen2.5-3B).}
    TSR modifies rollout construction, while GiGPO provides finer-grained credit assignment.
    Success rates are reported as mean $\pm$ standard deviation across runs.}
    \label{tab:gigpo_tsr_composition}
    \resizebox{0.5\linewidth}{!}{%
    \begin{tabular}{lcc}
        \toprule
        \textbf{Method}
        & \textbf{FrozenLake}
        & \textbf{WebShop} \\
        \midrule
        GiGPO
        & $56.8 \pm 1.5$
        & $80.2 \pm 1.3$ \\

        TSR (Beam Search)
        & $60.7 \pm 1.2$
        & $85.3 \pm 1.1$ \\

        \rowcolor{neugreenpastel!30}
        GiGPO + TSR
        & $\mathbf{62.4 \pm 1.2}$
        & $\mathbf{87.0 \pm 1.0}$ \\
        \bottomrule
    \end{tabular}
    }
\end{table}

\paragraph{Summary.}
These comparisons show that TSR remains effective against strong multi-turn RL baselines rather than only against naive rollout sampling.
TSR achieves higher performance than both ARPO and GiGPO under the matched training setup.
Its advantage over ARPO persists when both methods receive identical rollout-scoring information, directly separating the contribution of the rollout-search mechanism from that of the scoring signal.
Moreover, combining TSR with GiGPO yields further improvements over either method alone, showing that structured trajectory search is complementary to finer-grained credit assignment.
Overall, these results establish TSR as a distinct and composable mechanism for improving rollout generation in multi-turn RL.

\subsection{Optimizer Compatibility}
\label{app:optimizer_compatibility}
TSR modifies only the rollout-generation procedure and does not introduce an optimizer-specific loss, advantage estimator, or policy-update rule.
We therefore describe TSR as \emph{optimizer-compatible by design}.
To verify that its empirical gains are not specific to the optimizer choices used in our main experiments, we perform a controlled comparison on Sokoban with Qwen2.5-3B using both PPO and GRPO.
For each optimizer, we keep the model, task, rollout configuration, scoring function, and evaluation protocol fixed, and compare instance filtering against TSR (Beam Search).
As shown in Table~\ref{tab:optimizer_compatibility}, TSR improves performance under both optimization methods.
With PPO, TSR increases success from $43.7\%$ to $52.3\%$, corresponding to an $8.6$ pp improvement.
With GRPO, success increases from $42.4\%$ to $50.1\%$, corresponding to a $7.7$ pp improvement.
TSR also reduces average response length and the number of interaction turns under both optimizers.
These consistent gains support the view that the benefit of search-guided rollout generation is not specific to either PPO or GRPO.
Importantly, we do not claim that optimizer choice is irrelevant to absolute performance.
Rather, TSR can be integrated with different policy-gradient optimizers without changing their respective update rules, and the rollout-search benefit persists across both optimizers considered here.

\begin{table}[htbp]
    \centering
    \caption{\textbf{Optimizer Compatibility on Sokoban (Qwen2.5-3B).}
    We compare instance filtering and TSR (Beam Search) under PPO and GRPO while keeping the model, rollout, scoring, and evaluation configurations fixed. Success rates are reported as mean $\pm$ standard deviation across three runs.}
    \label{tab:optimizer_compatibility}
    \resizebox{0.8\linewidth}{!}{%
    \begin{tabular}{llccc}
        \toprule
        \textbf{Optimizer}
        & \textbf{Method}
        & \textbf{Success Rate ($\uparrow$)}
        & \textbf{Resp. Len. ($\downarrow$)}
        & \textbf{Turns ($\downarrow$)} \\
        \midrule

        \multirow{2}{*}{PPO}
        & Instance Filtering
        & $43.7 \pm 1.6$
        & 161
        & 4.1 \\

        & \cellcolor{neugreenpastel!30}TSR (Beam Search)
        & \cellcolor{neugreenpastel!30}$\mathbf{52.3 \pm 1.3}$
        & \cellcolor{neugreenpastel!30}\textbf{152}
        & \cellcolor{neugreenpastel!30}\textbf{3.6} \\

        \midrule

        \multirow{2}{*}{GRPO}
        & Instance Filtering
        & $42.4 \pm 1.8$
        & 166
        & 4.2 \\

        & \cellcolor{neugreenpastel!30}TSR (Beam Search)
        & \cellcolor{neugreenpastel!30}$\mathbf{50.1 \pm 1.5}$
        & \cellcolor{neugreenpastel!30}\textbf{156}
        & \cellcolor{neugreenpastel!30}\textbf{3.7} \\

        \bottomrule
    \end{tabular}
    }
\end{table}

\subsection{Comparison to Larger Zero-Shot Models}
\label{app:supplementary_results_larger_models}
Consistent with multi-turn agent RL evaluations in prior works~\citep{wang2025ragenunderstandingselfevolutionllm}, we additionally compare smaller TSR-trained Qwen2.5-0.5B agents against larger zero-shot models to contextualize the magnitude of the task-specific RL gains.
Importantly, this comparison is not intended as a compute- or training-matched evaluation as the larger models are used without task-specific RL, whereas the agents are trained on the target environments.
Instead, the goal is to illustrate that improving rollout generation during multi-turn RL can substantially close the performance gap to much larger generalist models, and in some cases exceed their zero-shot performance on specialized agentic tasks.
As shown in Table~\ref{tab:comparison_other_models}, task-specific RL substantially improves the smaller Qwen2.5-0.5B agent model performance.
With TSR (Beam Search), the 0.5B model reaches $38.30\%$ success on Sokoban and $30.00\%$ on FrozenLake, exceeding the zero-shot GPT-4o and Qwen2.5-72B baselines in this evaluation.
We emphasize that this does not imply that the smaller TSR-trained model is generally stronger than these larger models.
Rather, the result highlights that multi-turn RL with improved rollout generation can produce strong specialized agents, even from comparatively small backbones.
This supports our broader conclusion that train-time trajectory search is an effective mechanism for converting additional training compute into improved agentic task performance.

\begin{table}[htbp]
    \centering
    \caption{\textbf{Comparison to Larger Zero-Shot Models.} This comparison contextualizes the magnitude of task-specific RL gains and is not intended as a compute- or training-matched comparison.}
    \label{tab:comparison_other_models}
    \resizebox{0.6\linewidth}{!}{%
    \begin{tabular}{lcc}
        \toprule
        \textbf{Model / Method} & \textbf{Sokoban} & \textbf{FrozenLake} \\
        \midrule
        GPT-4o (zero-shot)              & 27.73 & 26.56 \\
        Qwen2.5-72B (zero-shot)         & 19.53 & 23.83 \\
        \midrule
        Qwen2.5-0.5B (Instance Filtering)  & 29.00 & 19.70 \\
        Qwen2.5-0.5B (TSR Best-of-$N$)     & 33.30 & 25.00 \\
        Qwen2.5-0.5B (TSR Lookahead)       & 36.10 & 27.80 \\
        \rowcolor{neugreenpastel!30}
        Qwen2.5-0.5B (TSR Beam Search)     & \textbf{38.30} & \textbf{30.00} \\
        \bottomrule
    \end{tabular}
    }
\end{table}

\begin{table}[htbp]
    \centering
    \caption{\textbf{Additional Agentic Environments.}
    Results on ALFWorld and AppWorld for Qwen2.5-3B and Qwen2.5-7B.
    We report ALFWorld task success rate and AppWorld Task Goal Completion (TGC).
    Results are mean $\pm$ standard deviation across three runs.}
    \label{tab:additional_agentic_environments}
    \resizebox{0.78\linewidth}{!}{%
    \begin{tabular}{llcc}
        \toprule
        \textbf{Model}
        & \textbf{Method}
        & \textbf{ALFWorld SR ($\uparrow$)}
        & \textbf{AppWorld TGC ($\uparrow$)} \\
        \midrule

        \multirow{4}{*}{Qwen2.5-3B}
        & Instance Filtering
        & $69.8 \pm 1.8$
        & $20.4 \pm 1.6$ \\

        & ARPO
        & $74.6 \pm 1.6$
        & $23.6 \pm 1.5$ \\

        & GiGPO
        & $78.3 \pm 1.4$
        & $25.2 \pm 1.4$ \\

        \rowcolor{neugreenpastel!30}
        \cellcolor{white}
        & TSR (Beam Search)
        & $\mathbf{84.7 \pm 1.3}$
        & $\mathbf{31.5 \pm 1.2}$ \\

        \midrule

        \multirow{4}{*}{Qwen2.5-7B}
        & Instance Filtering
        & $79.4 \pm 1.5$
        & $28.7 \pm 1.5$ \\

        & ARPO
        & $84.1 \pm 1.4$
        & $32.1 \pm 1.4$ \\

        & GiGPO
        & $89.3 \pm 1.2$
        & $34.8 \pm 1.3$ \\

        \rowcolor{neugreenpastel!30}
        \cellcolor{white}
        & TSR (Beam Search)
        & $\mathbf{92.1 \pm 1.0}$
        & $\mathbf{40.6 \pm 1.1}$ \\

        \bottomrule
    \end{tabular}
    }
\end{table}

\subsection{Additional Agentic Environments}
\label{app:additional_agentic_environments}
To evaluate whether the benefits of TSR extend beyond the three environments considered in the main experiments, we additionally evaluate TSR on ALFWorld~\citep{shridhar2021alfworldaligningtextembodied} and AppWorld~\citep{trivedi2024appworldcontrollableworldapps}.
These environments broaden the evaluation to two complementary agentic settings.
ALFWorld evaluates long-horizon household interaction and planning through text-based environments aligned with embodied tasks, while AppWorld evaluates interactive tool use and coding across multiple simulated applications and APIs.
We evaluate both Qwen2.5-3B-Instruct and the larger Qwen2.5-7B-Instruct model to additionally test whether the benefits of trajectory search persist at a larger model scale.
For each environment and model size, we follow the original benchmark task definitions and evaluation protocols, while all compared methods use the same task distribution, observation and action interfaces, policy optimizer, policy-optimization reward, and rollout budget.
We compare TSR (Beam Search) against instance filtering, ARPO, and GiGPO.
For ALFWorld, we train on the standard training games and report task success rate on the unseen evaluation split.
For AppWorld, we train on the train split and evaluate on test-normal, reporting Task Goal Completion (TGC), which counts a task as completed only when all corresponding evaluation tests pass.
In both environments, TSR ranks candidate prefixes using an observation-only value model trained analogously to~\refapp{app:disentangling_search_from_scoring_signals}, i.e., on baseline-generated trajectories to predict eventual task completion from the agent's observations and interaction history.
All scores are averaged across three independent runs.
Table~\ref{tab:additional_agentic_environments} summarizes the results.

\paragraph{Branching and Environment State.}
Branching requires restoring previously visited environment states.
We checkpoint the full simulator state at each branch point, which for AppWorld includes the application databases and the agent's interpreter state.
The same snapshot-restore interface is used for all methods that branch, and all transitions generated during branching count toward the matched rollout-generation budget.
TSR therefore receives no additional task information but requires a checkpointable training environment, which we treat as an explicit applicability condition.

\paragraph{Performance.}
Across both environments and model scales, TSR achieves the highest mean task performance.
For Qwen2.5-3B, TSR reaches $84.7\%$ success on ALFWorld compared with $69.8\%$ for instance filtering, $74.6\%$ for ARPO, and $78.3\%$ for GiGPO.
On AppWorld, TSR achieves $31.5\%$ TGC, compared with $20.4\%$ for instance filtering, $23.6\%$ for ARPO, and $25.2\%$ for GiGPO.
This corresponds to gains of $14.9$ percentage points on ALFWorld and $11.1$ points on AppWorld over instance filtering.
The same trend persists with Qwen2.5-7B.
TSR achieves $92.1\%$ success on ALFWorld and $40.6\%$ TGC on AppWorld, compared with $79.4\%$ and $28.7\%$ for instance filtering.
TSR also outperforms both ARPO and GiGPO at the larger model scale, reaching the strongest performance across both additional environments.
The corresponding improvements over instance filtering are $12.7$ percentage points on ALFWorld and $11.9$ points on AppWorld.
These results show that the gains from search-guided rollout generation extend beyond the original planning, navigation, and web-interaction environments.

\paragraph{Model-Scale Robustness.}
Increasing the backbone size improves absolute performance for all methods without diminishing the benefit of structured rollout search.
For TSR, scaling from Qwen2.5-3B to Qwen2.5-7B improves ALFWorld success from $84.7\%$ to $92.1\%$ and AppWorld TGC from $31.5\%$ to $40.6\%$.
Importantly, TSR maintains gains of approximately $11$--$15$ percentage points over instance filtering across both environments and model scales.
Its advantage over ARPO and GiGPO also persists at 7B, providing evidence that the benefits of trajectory-search rollouts are not confined to the smaller backbone used in the main experiments.

\paragraph{Summary.}
The additional evaluations extend the empirical scope of TSR along two complementary dimensions.
First, TSR improves performance on household interaction and multi-application tool-use tasks that differ substantially from Sokoban, FrozenLake, and WebShop.
Second, the improvement persists when scaling the policy from Qwen2.5-3B to Qwen2.5-7B.
Together with the main experiments, these results provide evidence that structured train-time trajectory search is effective across a broader range of multi-turn agentic environments and model scales.

%% file: appendix/G_limitations.tex
\section{Limitations and Broader Impact}
\label{app:limitations}

\paragraph{Limitations.}
TSR introduces additional compute during rollout generation, although this cost is incurred only at training time and is partially offset by requiring fewer optimization steps to convergence.
TSR also assumes that the environment can be branched, restored, or re-queried at intermediate states, which holds naturally for the digital and simulator-based environments common in LLM-agent research (e.g., web sandboxes, text games, and code interpreters), but may require learned environment models in non-resettable settings.
Our focus is therefore on resettable or simulatable training environments, while extending TSR to less controllable real-world settings remains important future work.
Furthermore, when immediate environment rewards are sparse, more informative prefix scoring can improve search efficiency.
In our main experiments, we use simple state-/progress-based signals (e.g., Manhattan distance and keyword overlap), while less structured environments may instead benefit from learned value or process-reward models.
Importantly, our controls show that TSR remains effective with task rewards, task-agnostic signals, learned observation-only values, noisy scores, and Monte Carlo-backed terminal rewards, demonstrating that it is not tied to a particular scoring mechanism.
Finally, our main evaluation spans two model scales and three complementary environments, with additional experiments extending the study to ALFWorld and AppWorld and to a 7B model.
Evaluating further model families and interaction settings remains an important direction for assessing the broader applicability of TSR.

\paragraph{Broader Impact.}
TSR changes only training-time rollout generation and does not introduce additional deployment-time components or actions.
By improving rollout quality, it can produce stronger policies with fewer interaction turns at inference time without requiring search at deployment, although this may come at additional training-time compute.
Additionally, by enabling comparatively smaller models to achieve stronger task-specific performance, TSR may also lower the computational barrier to multi-turn agent research.
As with other methods that improve agent capabilities, deployments involving external tools, transactions, code execution, or user data should retain appropriate safeguards and access controls.

\paragraph{Contributions.}
This work presents a systematic study of search-guided rollout generation for multi-turn RL of LLM agents.
We introduce TSR, a policy-proximal and optimizer-compatible framework that moves structured trajectory search to training-time rollouts and distills the resulting behavior into the policy.
Across five agentic environments and model scales from Qwen2.5-0.5B to Qwen2.5-7B, TSR consistently improves task performance while requiring no search at inference time.
Our evaluation further isolates the contribution of structured search through rollout-budget-matched baselines, information-matched scoring controls, Monte Carlo-backed search, optimizer comparisons, evaluations against related multi-turn RL baselines, and extensive analyses of convergence, compute, and distribution shift.
Together, these results establish TSR as a modular and empirically validated mechanism for improving rollout generation in multi-turn agent learning.